\documentclass[11pt,a4paper]{article}
\usepackage{times,latexsym}
\usepackage{url}
\usepackage[T1]{fontenc}
\usepackage{caption}
\usepackage{subcaption}

\usepackage[acceptedWithA,]{tacl2021v1}
\usepackage{graphicx}
\usepackage{booktabs}
\usepackage{threeparttable} 
\usepackage{xspace,mfirstuc,tabulary}

\newif\iftaclinstructions
\taclinstructionsfalse 
\iftaclinstructions
\renewcommand{\confidential}{}
\renewcommand{\anonsubtext}{(No author info supplied here, for consistency with
TACL-submission anonymization requirements)}
\newcommand{\instr}
\fi

\iftaclpubformat 

\else

\fi

 \definecolor{MidnightBlue}{rgb}{0,0.403,0.58}

\newcommand\todo[1]{\textcolor{black}{#1}}
\newcommand\fromApp[1]{\textcolor{black}{#1}}

\newcommand{\bestRegAndSim}[1]{\textcolor{MidnightBlue}{\underline{#1}}}
\newcommand{\bestSim}[1]{\textcolor{MidnightBlue}{#1}}

\newcommand{\bestReg}[1]{\underline{#1}}

\newcommand\qqq[1]{\textcolor{gray}{#1}}

\usepackage{amsmath}

\title{Human Choice Prediction in Language-based Persuasion Games: Simulation-based Off-Policy Evaluation
}

\author{
    \textbf{Eilam Shapira} \quad \textbf{Omer Madmon} \quad \textbf{Reut Apel} \quad \textbf{Moshe Tennenholtz} \quad \textbf{Roi Reichart}\\
    Faculty of Data and Decision Sciences, Technion - Israel Institute of Technology \\
    \texttt{\{eilam.shapira, omermadmon, reutapel88,} \\ \texttt{moshe.tennenholtz, roireichart\}@gmail.com}}

\date{}

\begin{document}
\maketitle

\begingroup
\renewcommand\thefootnote{}\footnote{
Accepted for publication in Transactions of the Association for Computational Linguistics (TACL), 2025. Pre-MIT Press publication version.}
\addtocounter{footnote}{-1}
\endgroup

\begin{abstract}
Recent advances in Large Language Models (LLMs) have spurred interest in designing LLM-based agents for tasks that involve interaction with human and artificial agents. This paper addresses a key aspect in the design of such agents: predicting human decisions in off-policy evaluation (OPE).
We focus on language-based persuasion games, where an expert aims to influence the decision-maker through verbal messages.
In our OPE framework, the prediction model is trained on human interaction data collected from encounters with one set of expert agents, and its performance is evaluated on interactions with a different set of experts.
Using a dedicated application, we collected a dataset of 87K decisions from humans playing a repeated decision-making game with artificial agents. 
To enhance off-policy performance, we propose a simulation technique involving interactions across the entire agent space and simulated decision-makers. Our learning strategy yields significant OPE gains, e.g., improving prediction accuracy in the top 15\% challenging cases by 7.1\%.\footnote{Our data and code are available in the GitHub repository: https://github.com/eilamshapira/HumanChoicePrediction}
\end{abstract}

\section{Introduction}
\label{intro}


Consider an online platform like Booking.com, where service providers (e.g., hotel owners) promote their services to potential consumers (e.g., travelers). These platforms enable various economic interactions with dynamic behavior, making reputation a key factor as the interaction is often repeated. The platform often aims to \emph{predict user behavior} with service providers for tasks like revenue forecasting and improved matching to boost social welfare. Predicting user behavior with new, unseen providers, however, results in a \emph{distribution shift}. 
In this paper, we introduce a novel approach to address this prediction challenge. We use the term \emph{Off-Policy Evaluation (OPE)} to describe a scenario where test-time interactions involve behavioral patterns and strategies from service providers that differ from those in the training data. When test-time interactions align with the training distribution, we refer to this as the \emph{on-policy} scenario.

The interaction described above can be modeled as a game with asymmetric information, famously known as a \emph{persuasion game}. In this game, a \emph{sender} (i.e., a hotel owner or a travel agent) aims to influence the decision of a \emph{receiver} (i.e., the consumer) through strategic communication. Unlike zero-sum games,\footnote{The term zero-sum game typically refers to a two-player game where one player’s gain comes at the expense of the other, implying a complete misalignment of interests—rarely seen in real-world interactions.} persuasion games may involve partially aligned or misaligned interests, depending on the \emph{state of the world} (hotel quality)—only observed by the sender.

Economics emphasizes the importance of studying non-cooperative games beyond zero-sum scenarios \cite{mas-colell.whinston.ea95}, with persuasion games being central to information economics \cite{aumann_maschler_stearns_1995,kamenica_gentzkow_2009,emek2014signaling,bahar2016economic,BergemannMorris}. However, many game-theoretic models rely on simplified messaging and overlook the complexities of natural language communication between senders and receivers.
Although their incentives may differ, they are not in complete opposition, making straightforward maximization solutions inadequate \cite{fudenberg1991game}. Unlike traditional economic models that use formal signals, we explore persuasion games with \emph{natural language communication}.


\begin{figure*}[t!]
  \centering
  \includegraphics[width=\textwidth]{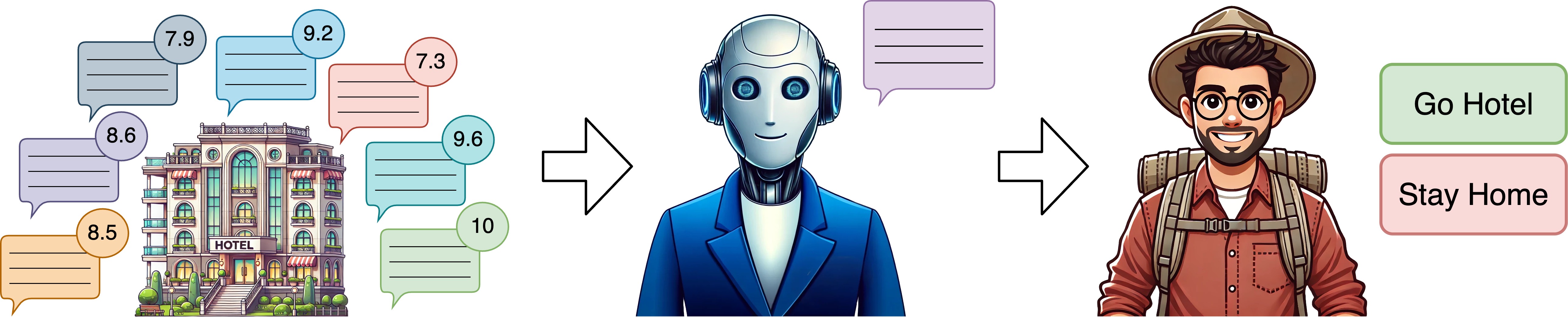}
  \caption{\todo{Illustration of a single round in the language-based persuasion game. The bot expert starts by analyzing the interaction history from prior rounds (not depicted in the illustration) alongside a set of seven reviews, each consisting of a textual description and an associated score. Following a predefined strategy, it selects one review from the set and transmits only its textual content to the human Decision Maker (DM). The DM then evaluates the received review in the context of the full interaction history and chooses an action. In the final step, both the expert and the DM receive their payoffs, which are determined by the DM’s choice and the hotel's actual quality.}}
    \label{fig:a_round}
\end{figure*}

Recent research \cite{apel2020predicting,meta_human-level_2022,raifer2022designing} has ventured into language-based games, showcasing notable success in playing them, with diplomacy, a multi-person zero-sum game, being a noteworthy example \cite{meta_human-level_2022}. Despite these strides, a crucial gap persists in our comprehension of human choice within non-cooperative language-based persuasion games.

\citet{apel2020predicting} introduced a unique non-cooperative language-based persuasion game, featuring a multi-stage setup involving an \textit{expert} (travel agent) and a \textit{decision-maker (DM)}, the customer. In each interaction, the expert selects a scored textual review from the hotel's reviews to persuade the decision-maker to choose the hotel. The DM's acceptance or rejection yields stochastic payoffs determined by the review score distribution, accessible only to the expert. Both players move to the next stage with a similar structure after observing payoffs, but with a different hotel. Figure \ref{fig:a_round} illustrates a single round of the game. While \citet{apel2020predicting} primarily focused on predicting DM actions, \citet{raifer2022designing} adapted the framework, creating an artificial expert (AE) employing the Monte Carlo Tree Search (MCTS) algorithm \citep{coulom2006efficient}. The AE utilizes deep learning models, incorporating behavioral and linguistic features to anticipate the DM's actions, and predict the expert's future reward based on game status and a potential review. The AE aims to maximize the number of hotels accepted by the DM.


\subsection {Our Contribution}
This paper focuses on \textit{off-policy human choice prediction in language-based persuasion games}. To assess the comprehensibility of human decisions, we consider the prediction of human behavior when faced with an \textit{unobserved opponent}. Instead of determining optimal policies, we aim to predict human agents' choices when playing with a set of artificial experts in a given game, based on their interactions with various other experts in the same game. This is an OPE setup for experts (agents) interacting with human decision-makers (DMs) in a persuasion game.

\paragraph{Data}
To realize this objective, we present a mobile application simulating a realistic language-based persuasion game environment. Through experiments involving human agents engaging with diverse artificial agents, we aim to establish predictive models that elucidate how humans respond to unfamiliar partners based on their interactions with known counterparts. Particularly, our dataset consists of 87k decisions from 245 DMs who played against 12 different automatic expert bots (each DM played against 6 bots). We consider this dataset as a contribution to the research community and will make it public, hoping that it will promote the research in our area.

\paragraph{Simulation}
To enhance the performance of human choice prediction in OPE, we take a \emph{simulation-based algorithmic approach}.
We address data constraints in modeling human interactions by combining human-bot and simulated DM-bot interactions. Our DM simulation model assumes that DMs utilize a combination of heuristics related to past game behavior of both players and the content of the chosen review, and that they improve over time regardless of the specific strategy of the expert. 
From an algorithmic perspective, the simulation is designed to model a DM that utilizes a \emph{mixture-of-heuristics with dynamic weights (probabilities)}, where the weight of an \emph{oracle heuristic} (a DM that knows the optimal decision) increases over time. This idea is inspired by the \emph{multiplicative weights} algorithm, commonly used in online learning, game theory, and optimization \cite{fudenberg1995consistency,freund1999adaptive,arora2012multiplicative}.
The game-agnostic improvement-over-time principle enables data generation from interactions between simulated DMs and diverse bots. Our results indicate that training a human decision prediction model on this mix of human interaction and simulated data results in a more robust model, not tailored to specific bot idiosyncrasies in the training set, making it suitable for predictions involving new bots and human DMs. Our Ablation analysis highlights the importance of the various components of our simulation: learning-over-time, past behaviour and review content modeling.

\section{Related Work}

\subsection{Persuasion in NLP}
Persuasion has been extensively explored in NLP throughout the years.
\citet{tan2016winning} contributed a vital dataset from Reddit's ChangeMyView for online persuasion analysis. \citet{hidey_analyzing_2017} explored argument classification in online persuasion, while \citet{hidey2018persuasive} examined the impact of argument sequencing on persuasive success. \citet{wang_persuasion_2019} investigated persuasive dialogue systems aimed at social good. \citet{yang2019let} developed predictive models that assess the persuasiveness of requests on crowdfunding platforms. \citet{chen2021weakly} offered a text repository for identifying effective persuasive strategies. \citet{hiraoka2014reinforcement} applied reinforcement learning to cooperative persuasive dialogues.

Several works focused on studying persuasion from the expert's perspective: \citet{raifer2022designing} followed the setup of \citet{apel2020predicting} to design an automated expert for language-based persuasion games, utilizing tools such as MCTS; \citet{carrascofarre2024largelanguagemodelspersuasive} study persuasive strategies employed by LLMs; \citet{breum2024persuasive} study the effect of persuasive LLMs on \emph{opinion dynamics}; and \citet{matz2024potential} demonstrate the potential of LLMs in \emph{personalized} persuasion.
In contrast, we focus on predicting the behavior of human \emph{decision-makers}, particularly in the \emph{off-policy evaluation} scenario, and developing a novel simulation-based approach.

\subsection{Simulation Data}
Simulation ideas have been flowering in ML areas where human-human and human-machine interactions are modeled, e.g., in Reinforcement Learning (RL), due to the costly and laborious data collection for such setups \cite{tesauro1991practical}. Notable applications include RL for robotics \citep{bousmalis2018using, vacaro2019sim}, and autonomous cars \citep{yue2018lidar}. 
Simulations also play a crucial role in the development of artificial agents proficient in gaming scenarios, e.g. by using MCTS-like simulations to enhance agent performance \cite{silveretal,SilverSSAHGHBLB17,SchrittwieserAH20,oroojlooy2022review}.
In NLP, simulating human interactions is used to build dialog systems \citep{jung_integrated_2008, ai2008user, gonzalez2010cooperative, shi_how_2019,zhang_evaluating_2020,liu2023training} and train LLM-based agents that mimic human behavior \cite{Park:23,hussain2023tutorial,chuang2023simulating,Taubenfeld:24}.

Our work demonstrates a novel use of integrating interaction data with simulation data. Doing this we step in the footpath of several works in diverse domains, such as NLP \citep{calderon2022docogen}, autonomous cars \citep{cao2022data, yue2018lidar}, and astro-particle physics \citep{saadallah2022simulation}. 
Our simulation is novel as it integrates simple heuristics and can shed light on human behavior. 
It is also designed to model a DM that, like human DMs, learns and improves over time--a property that is shown to be highly effective in enhancing OPE.

\begin{figure*}[t!]
  \centering
  \includegraphics[width=\textwidth]{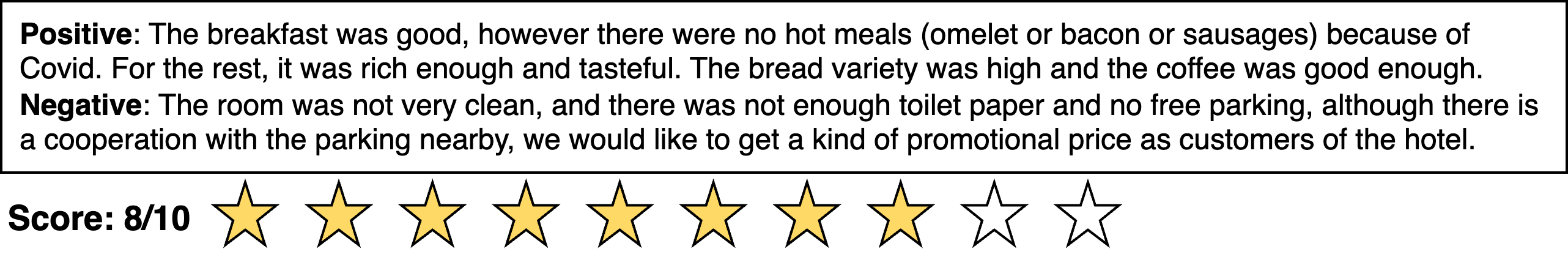}
  \caption{\fromApp{A sample review from our hotel review dataset. The agent is exposed to both the textual part and the numerical rating of the review. The agent sends only the textual signal to the DM, which is not exposed to the numerical rating.}}
    \label{fig:review}
\end{figure*}

\subsection{Action Prediction in ML and NLP}

In the realm of ML and NLP, action prediction, particularly in human decision-making, has been studied across diverse scenarios \cite{plonsky2019predicting, rosenfeld2018predicting, bourgin2019cognitive}. For example, \citet{psyforest_plonsky_2017} integrated psychological features with ML techniques, focusing on decision-making in games against nature, while \citet{auletta2023predicting} utilized supervised learning and explainable AI techniques for action prediction in collaborative tasks.
While these works have not involved language, others try to predict human decisions in language-based situations.
\citet{ben2020predicting} predicted individuals' actions in one-shot games based on free-text responses, and \citet{nadavAmirRoi2020} forecast the in-game actions of NBA players by leveraging insights from open-ended interviews. 

Language-based action prediction has been extensively explored in the legal domain: \citet{zhong2018legal} and \citet{yang2019recurrent} developed novel approaches for judgment prediction; \citet{bak_conversational_2018} demonstrated how group discussions can be used to predict a leader's decision; \citet{aletras_predicting_2016} and \citet{medvedeva2020using} utilized ML and NLP to predict decisions of the European Court of Human Rights.

The recent advancements in LLMs for strategic and economic scenarios have opened new possibilities for leveraging LLMs as data generators to predict human actions in economic environments \cite{xi2023rise}. For instance, \citet{horton2023large} studied the behavior of LLMs in well-known behavioral economics experiments; \citet{chen2023emergence} studied the emergence of rationality of GPT; \citet{akata2023playing,guo2024economics} compared the behavior of LLMs in games to those of rational agents, as predicted by game-theoretic concepts; and \citet{shapira2024glee} assessed efficiency and fairness of LLMs in games.

Closest to our work, \citet{shapira_can_2024} demonstrated the potential of this approach in a similar setting to our language-based persuasion game. While this LLM-based approach is promising, our simulation-based approach offers three key advantages: (a) it is significantly more cost-effective, both in terms of budget and runtime; (b) it proves effective in the OPE setting, which was not studied by \citet{shapira_can_2024};  and (c) it serves as an interpretable generative model for human choice decisions.
\section{Problem Definition}
While the space of non-cooperative games is very large, our emphasis here is on language-based persuasion games, in which textual messages replace the stylized messages discussed in economic theory. These games model interactions that typically arise in real-world applications such as online platforms, as illustrated in \S\ref{intro}.

\paragraph{Language-based Persuasion Game}
The game consists of two parties, an {\em expert} and a {\em decision-maker (DM)}, interacting for \textit{R} rounds. In each round, the expert, who plays the role of a travel agent, attempts to promote a randomly selected hotel. The expert is presented with $m$ scored reviews that were written and scored by real users of Booking.com. The expert is then asked to send the DM one of the reviews to persuade her to select the hotel. Figure \ref{fig:review} presents an example review.
A hotel is considered good if: 

\begin{equation}
\label{hotel-quality-eq}
\hat{s} = \frac{1}{m}\sum\limits_{i=1}^{m} s_i \geq TH
\end{equation}

i.e. its average review score, $\hat{s}$, is not less than a predefined threshold, $TH$, and bad otherwise, where $s_i$ is the hotel's $i'th$  review score.  

In the experimental study (see next section), following \citet{apel2020predicting}, we take \textit{R = 10} and \textit{$m$ = 7}.  
We chose \textit{TH = 8.0} because, according to Booking.com, a hotel rated 8.0 or higher is considered a good hotel. The definition of what constitutes a good hotel is available only to the expert.

While the expert observes both the verbal and the numerical part of the reviews and hence knows the hotel's quality, the DM observes only the verbal part of the review sent to her.
The DM's task is to decide whether to accept the expert's offer and go to the hotel or decline it and stay at home, based only on the review provided by the expert. The DM's payoff at each round depends on the quality of the hotel, with a positive payoff (of 1 point) received when a good hotel is selected or when a bad hotel is not selected, and a payoff of 0 incurred otherwise. At the end of each round, both players are notified of the DM's decision and her individual payoff. The goal of the DM is to gain at least $TR$ out of the $R=10$ possible points (see \S\ref{sec:data_collection}).

\begin{table*}[htp]
  \centering
  \begin{tabular}{ll}
  
    \hline
Split Condition Description  & Condition formulation \\
    \hline
Is the current hotel good?  &  $\hat{s}_t \ge 8$ \\
Did the DM choose to go to the hotel in the previous round?  &   $d_{t-1} = 1$ \\
Was the hotel in the previous round good?  &   $\hat{s}_{t-1} \ge 8$  \\
Has the decision maker earned more points than the &\\number of times he chose to go to the hotels?
 & $\sum_{i=1}^{t-1} (I_{\hat{s}_{i} \ge 8}  = d_i) > \sum_{i=1}^{t-1} d_i $ \\
    \hline
Action Description  \\
    \hline
Send the $r$ review  &  $r \in$ \{best, mean, worst\} \\
    \hline
  \end{tabular}
    \caption{The conditions and actions employed by the rule-based experts in round $t$. The hotel score in the $i$-th round is denoted with $\hat{s}_i$, while the binary decision made by the decision maker in the $i$-th round is denoted with $d_i$ (with $d_i = 1$ for hotel selection).
  }
\label{bots-strategies}
\end{table*}

\subsection{Strategy Space} 

While \citet{apel2020predicting} study human vs. human games, in this work we generalized their game to human (DM) vs. bot (expert) interactions in an OPE setup. We therefore need to define a \emph{strategy space} for the experts. This space encompasses all simple, deterministic decision-tree-based strategies that can be constructed using a pre-defined set of binary split conditions and a pre-defined set of actions. These conditions and actions are based on the respective reviewers' numerical scores assigned to the hotels, and the game's history, as detailed in Table \ref{bots-strategies}. Employing decision trees of depth up to 2 (to keep the strategies simple), we obtained a total of 1179 strategies. 

Six of these strategies were selected for group $E_\mathcal{A}$, and six others for group $E_\mathcal{B}$, each of the groups is played with a different set of human DMs to implement an OPE setting. The strategies were selected so that they are different from each other, and the difference between $E_\mathcal{A}$ and $E_\mathcal{B}$ is large.\footnote{To understand what 'difference' between strategies means, notice that each strategy induces a probability distribution over the review scores it reveals to the DM. Then, similarity can be naturally defined between any pair of such induced distributions. As we discuss next, our OPE task is more challenging than its on-policy counterpart. This observation confirms that the selection process successfully identified two conceptually different strategy sets.} One example of such a strategy is presented in Figure \ref{fig:a-bot}.  A full list of the strategies in $E_\mathcal{A}$ and $E_\mathcal{B}$ is in Appendix \ref{appendix:strategies}.

\begin{figure}[htp]
  \centering
    \includegraphics[bb=0 0 500 400, width=0.7\columnwidth,trim={1cm 1cm 1cm 2.1cm},clip]{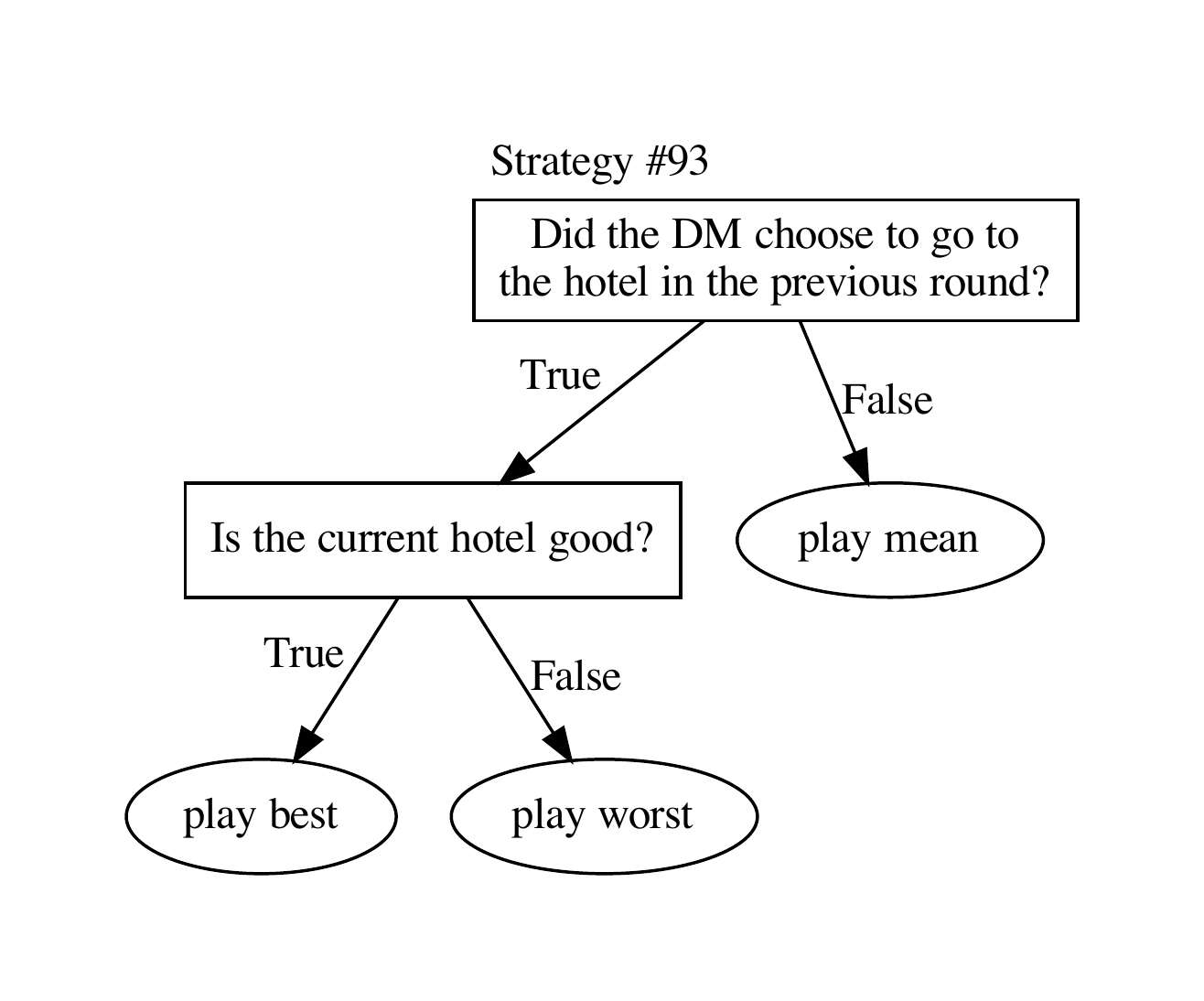}
  \caption{An example strategy from the $E_\mathcal{B}$ set.}
\label{fig:a-bot}
\end{figure}

An important aspect of the selected expert strategies is that they are simple and intuitive, and represent a diverse set of behavioral patterns that are likely to arise in real-world persuasion scenarios. Simplicity mostly follows from the fact that decision trees are restricted to depth 2. This property aligns with humans' tendency to follow simple heuristics due to limited cognitive resources \cite{hutchinson2005simple,gigerenzer2009homo}. To demonstrate the intuitive nature of these strategies, we first provide some examples of concrete strategies that are contained in $E_\mathcal{A}$ and $E_\mathcal{B}$, and then provide a high-level classification of the entire strategy space to conceptual categories, which are all represented in both strategy sets.

\paragraph{Examples}
We now introduce several intuitive persuasive behavioral patterns captured within particular expert strategies in our setup.
\begin{itemize}
    \item \emph{The greedy expert} always reveals the most optimistic piece of evidence for the hotel's quality, regardless of its true quality.
    \item \emph{The honest expert}  reveals the best review when the hotel is good, and reveals the most negative review when the true quality is low.
    \item \emph{The backward-looking expert} takes a different approach: when the previous decision of the DM was to accept the offer, she presents the best possible review, exploiting the good reputation to maintain momentum. Otherwise, she presents the (closest to the) mean-scored review, to avoid a bad reputation.
\end{itemize}

These experts take different persuasive approaches while having the same goal of maximizing their cumulative gain against a human DM. These reflect differing underlying beliefs about the human DMs behavior, and their performance can significantly differ depending on the opponent. The \emph{greedy} and \emph{honest} strategies are contained in $E_\mathcal{A}$, while the \emph{backward-looking} strategy is in $E_\mathcal{B}$.

\paragraph{Classification of Strategies}
We begin by observing that there are two types of split conditions according to which the strategies are constructed (Table \ref{bots-strategies}): conditions that depend on the \emph{hotel quality} (e.g., first row) and conditions that depend on the \emph{DM past behavior} (e.g., second row). A strategy may include both condition types, just one, or neither. It is therefore convenient to use this fact to define four distinct groups of strategies: \emph{(1) simple} strategies that include no split condition, and are defined solely by an action description (e.g., the \emph{greedy} strategy); \emph{(2) quality-dependent} that only contain split conditions depending on the hotel quality (e.g., the \emph{honest} strategy); \emph{(3) history-dependent} strategies, that contain only split conditions that are history dependent (e.g., the \emph{backward-looking} strategy); and \emph{(4) complex strategies} that contain both types of splitting conditions (e.g., the strategy illustrated in Figure \ref{fig:a-bot}). Importantly, our selected strategy sets are representative of the entire strategy space in the sense that they cover all four strategy classes.

\paragraph{Our Challenge} Given a dataset comprising of interactions between human decision-makers and an ordered set of experts $E_\mathcal{A}$,  our objective is to predict the behavior of other human decision-makers  when they engage in game-play with another ordered set of rule-based experts $E_\mathcal{B}$.\footnote{In Appendix \ref{app:ope} we demonstrate that the off-policy task is indeed harder than the on-policy one.}

\section{The Human-Bot Interaction Dataset}
\label{sec:data_collection}
In order to collect data, we  developed a mobile phone game application that follows the above multi-stage language-based persuasion game setting. In our game, a human DM plays with a series of 6 rule-based experts (bots, either $E_\mathcal{A}$ or $E_\mathcal{B}$), each game consisting of $R = 10$ rounds. The DM gets 1 point if she makes a good decision (selecting a good hotel or avoiding a bad one), and 0 points otherwise, and hence the maximal payoff is 10. To advance to the next level (play with the next bot), the DM must achieve a pre-defined target payoff. The target payoffs are in the 8-10 range, and are defined according to how challenging the bot is.\footnote{Based on game design considerations, we did not order the bots by difficulty; The target payoffs were estimated by the authors after playing several times against each bot.} 
The goal of the human player is to get the target payoff of all six experts. We refer to reaching the target payoff as "defeating" the expert, although this is not a zero-sum game with adversarial experts.%
\footnote{The introduction of target payoffs, whose goal is to enhance player engagement, represents yet another distinction of our work from that of \citet{apel2020predicting}.}

\paragraph{The Hotels} Utilizing hotel reviews sourced from Booking.com, we compiled a dataset comprising 1,068 hotels, each with $m=7$ scored reviews.
We chose the hotels so that only about half of them are defined as good (i.e., $\hat{s} \ge TH=8$). The median score of the hotels was also set to 8.01.

\paragraph{Interaction Data}

We ran our game in the Apple's App Store and Google Play for a few months (May 2022 - January 2023). The players who downloaded the app until November 2022 played with group $E_\mathcal{A}$ experts, while the players who played from December 2022 played with group $E_\mathcal{B}$ experts. We collected 87,204 decisions taken by 245 players who finished the game, i.e., defeated all six experts. Statistical details of the data are given in Table \ref{tab:data-statistics}.
We used reward schemes, including lottery participation and course credit, to incentivize players to beat all six experts in the game. More details about our app and the data collection process are in Appendix \ref{appendix:app}.

\begin{table*}[bhtp]
  \centering
  \begin{tabular}{llllll}
    \hline
Group & Experts & \#DMs  & \#decisions  & median \#decisions/DM & median \#games/DM \\
    \hline
A & $E_\mathcal{A}$     & 210                         & 71,579        & 273                 & 34.5                    \\
B & $E_\mathcal{B}$    & 35                          & 15,625        & 367               & 55                    \\
    \hline
Total & $E_\mathcal{A}$ or $E_\mathcal{B}$    & 245                          & 87,204        & 280               & 37                    \\
\hline
  \end{tabular}
  \caption{Dataset statistics.}
  \label{tab:data-statistics}
\end{table*}

\section{Simulation-based OPE}
\label{sec:simulation}

 We propose a simulation-based DM as an \emph{interpretable generative model for human choice data}. By interacting with expert bots using random strategies from the strategy space, the simulated DM generates data that is combined with initial human-bot interaction data to enhance off-policy prediction.\footnote{In Appendix \ref{app:onpolicy with sim}, we show that the simulation also improves on-policy prediction quality.} Algorithmically, the simulation uses a \emph{mixture of heuristics with dynamic weights}, relying on intuitive decision rules informed by past interactions and textual content to decide the next action.

Over time, the simulated DM dynamically updates the weights of these heuristics, referred to as its \emph{temperament}. However, while these heuristics are interpretable and intuitive, they are inherently simplistic and fail to emulate the adaptive learning seen in human DMs, whose performance improves over time \todo{(see \S \ref{app:human learning} for evidence)}.

To address this, we introduce an \emph{oracle} heuristic into the simulation, with its weight increasing over time. This adjustment enables the simulated DM to exhibit improvement patterns akin to human DMs, effectively combining human-like heuristics with gradual improvement. This approach proves highly effective in enhancing off-policy evaluation.




\paragraph{The Simulation}

\begin{figure*}[t!]
  \centering
  \includegraphics[width=\textwidth]{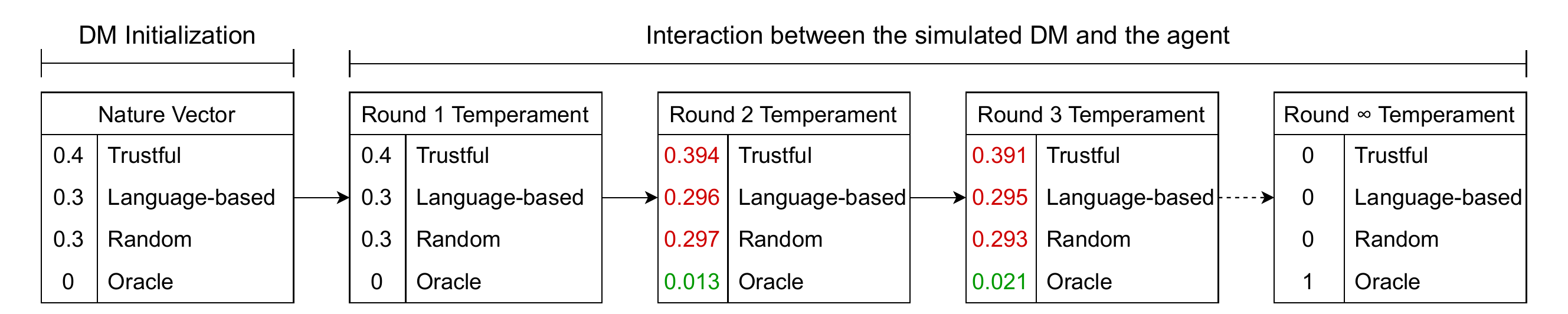}
  \caption{\todo{Example of the update process of the temperament vector of a simulated-DM.
  Each simulated DM is assigned a \textit{nature vector}, representing its inherent action probabilities. At the start of an interaction with a new agent, the DM’s \textit{temperament vector} is initialized to that nature vector. In each round, the DM’s action is randomly chosen according to the probabilities in the temperament vector. After the round, the temperament vector is updated so that, with some positive probability, the likelihood of playing Oracle increases, while the probabilities of playing all other actions decrease.}}
    \label{fig:temperament_vector}
\end{figure*}

In each instance of the bot-DM interaction simulation, we sample six expert strategies from the entire strategy space, uniformly at random. For each simulated DM-bot interaction, we randomly sample 10 hotels, one for each of the $R=10$ rounds. In each of the rounds, the expert uses its strategy to select a review from the review set of the hotel associated with that round. 
The simulation involves the DM playing the 10 rounds game against the same expert until achieving a payoff of $SIM\_PAY\_TH=9$ points, before moving on to play against the next expert.\footnote{If the DM does not reach a payoff of 9 or 10 then it repeats the game after 10 new hotels are sampled to replace the original 10 hotels.}  The simulated DM uses the textual review and its estimated numerical score (see below) to make decisions. 

Our simulation is based on two basic probability vectors: (a) The {\em nature vector}, a hyper-parameter vector denoted with $(p_1, p_2, p_3)$; This vector provides the initial probabilities that the DM will select one of three basic heuristics (see below); and (b) The {\em temperament vector}, comprised of four values $(p_0^t, p_1^t, p_2^t, p_3^t)$ and updated in each round $t$. While $p_1^t, p_2^t, p_3^t$ correspond to the three values in the nature vector, $p_0^t$ is the probability that the DM will play \textit{oracle}, and take the right decision just because it has learned how to play the game from the multi-stage interaction with the bot.
\footnote{A core idea behind the simulation design is that human DMs indeed learn and improve over time. In Appendix \ref{app:human learning} we provide empirical evidence of this phenomenon.}

The nature vector corresponds to three heuristics: \textit{Trustful, Language-based, and Random}. These heuristics reflect the two basic components we attribute to a DM: considering past behavior and its outcome (\textit{Trustful}) and learning from the information in the current hotel's review (\textit{Language-based}), alongside inherent randomness (\textit{Random}).\footnote{Notice that both the expert strategies and the structure of our simulation reflect underlying beliefs about the nature of human DMs. However, the simulation remains agnostic to the beliefs guiding the test-time strategies, and still produces high-quality data. In some sense, the simulation is hence based on heuristics that can be seen as fundamental.}

\paragraph{DM Heuristics}

Under the \textit{Trustful} heuristic, the DM chooses to go to the hotel if and only if in the last $K$ rounds the DM's estimated review score matched the feedback about the hotel quality, where $K$ is a stochastic parameter sampled for each DM individually. Notice that as opposed to the real human-bot interactions, in this heuristic the DM does use the numerical score of the review. However, in order to emulate the reality where it is hard for humans to accurately estimate the review score from its text, the estimated numerical score is defined as $\hat{s} + x$, where $\hat{s}$ is the actual score of the review and $x ~ \sim Normal(0, \epsilon)$ is a noise variable.
The hotel quality feedback is the average scores of the hotel's reviews (Equation \ref{hotel-quality-eq}).
According to the \textit{Language-based} heuristic, the DM uses an LLM to predict the review score. If the predicted score is 8 or higher, the DM chooses to go to the hotel. We computed review scores with Text-Bison \cite{anil_palm_2023}, prompting it to score each review on a 1-100 scale such that a good hotel is one with a score of $\geq 80$, and then re-scaled the scores into the 1-10 range. Finally, under the \textit{Random} heuristic, the DM would make a random decision.\footnote{Note that simulating data based on both textual and behavioral contexts using an LLM is financially prohibitive, as it would require us a call to the LLM for every decision in the simulation, for a total of tens to hundreds thousands calls.} 

\paragraph{The Temperament}


The temperament vector is initialized at the onset of each 10-round DM-bot interaction to be $p^0 = (0,p_1,p_2,p_3)$,
where $(p_1, p_2, p_3)$, the nature vector, is a DM-specific hyper-parameter (see Appendix \ref{appendix:HPT}). At each round $t$, the temperament vector is updated by multiplying $p_{1\le i \le 3}^t$ by a factor of $1 - \gamma_i^t$
, where $\gamma_i^t \sim Uni(-\frac{\eta}{10}, \eta)$ and $\eta \in (0,1]$ is a hyper-parameter representing the {\em DM's improvement rate}.\footnote{We allow $\gamma_i^t$ to get negative values since it is possible that at some rounds the DM performance degrades.}
Accordingly, $p_0^t$ is updated to be 
$
p_0^t = 1 - \sum_{1 \le i \le 3} p_i^t
$
to ensure that the temperament vector is a probability vector. 
In this way, the temperament vector after $T>0$ rounds is defined by:  
\begin{equation}
\label{eq:sim}
p_{i}^T = p_i \prod_{t=1}^{T}(1-\gamma_i^t)
\quad
\mathrm{and}
\quad 
p_0^T = 1 - \sum_{i=1}^3 p_i^T
\end{equation}


Since $0 < E [1 - \gamma_i^t] < 1$,  it holds that the probability making the right decision ($p_0^T$), irrespective of the nature vector,  tends towards 1 as the number of rounds approaches infinity. Hence, the DM will inevitably defeat any expert after a sufficient number of rounds.\footnote{In Eq. \ref{eq:sim}, it may be that $\sum_{1 \le i \le 3} p_i^T > 1$, in which case we trim this sum to 1 before computing $p_0^T $.} Figure \ref{fig:temperament_vector} illustrates the update process of the temperament vector.


\paragraph{Gradient-based Training} We leverage both simulation data and real human-bot interactions to train the decision prediction model. At the beginning of each training epoch, we train the model using $S_r$ simulated DMs per each human DM, and subsequently train the model using the human-bot interaction data ($S_r$ is a hyper-parameter). 






\section{Experiments}

\subsection{Feature Representation} 
\label{sec:HCF}

We represent each DM-bot interaction round with features related to (1) the hotel review sent by the expert to the DM; and (2) the strategic situation in which the decision was made.
To represent a review, we utilize a set of binary Engineered Features (EFs) originally proposed by \citet{apel2020predicting}.
These features describe the topics that the positive and negative parts of the review discuss (for example: Are the hotel's design mentioned in the positive part of the review?) as well as structural and stylistic properties of the review (e.g., Is the positive part shorter than the negative part?). 

To label the topics the review discusses, we use  OpenAI's Davinci model.\footnote{This stands in contrast to \citet{apel2020predicting}, who manually tagged their reviews with the EFs.} The model receives as a prompt the review and the feature definition, and is asked to indicate whether or not the feature appears in the review. Below we demonstrate that the use of EFs yielded better results compared to deep learning based text embedding techniques such as BERT \citep{kenton2019bert} and GPT-4 \citep{openai_gpt-4_2023}. 
To represent the strategic interaction, we introduce additional binary features that capture the DM's previous decision and outcome, current payoff, and frequency of choosing to go to the hotel in past rounds of the same interaction (see Appendix \ref{appendix:input} for further details).

\subsection{Models and Baselines}

This subsection provides a description of the models that we train for our study. For each model, except for the Majority Vote model, we train three versions: one using only human-bot interaction data, one using only simulation data, and one with both interaction and simulation data. This allows us to evaluate the impact of simulated data on the model's predictive performance. All the models are designed to predict the DM's decision in a specific round, given the previous rounds played in the same bot-DM interaction.

\paragraph{Majority Vote} 

The {\em Majority Vote} baseline predicts the DM's decision based on the percentage of DMs who decided to go to the hotel in the interaction training set. Notably, this baseline method solely relies on the review and disregards the repeated nature of the game. 
Additionally, it is unsuitable for predicting the decisions of players who are the first to encounter a new review. In order to make sure that we consider only cases where DMs indeed read the review, we consider only cases where DMs spent at least 3 seconds before making their decision.\footnote{Note that we could use this baseline because we use the same set of hotels at train, test (and also simulation) time. We justify this design choice by the large number of hotels in our dataset (1068, see \S\ref{sec:data_collection}), the resulting negligible probability of getting the same 10 hotel sequence in two different bot-DM interactions and the fact that in the actual world, the set of available hotels does not tend to change very quickly.}

\paragraph{Machine Learning Models}

We employ five machine learning models to predict the DM's decisions. First, we utilize a Long Short-Term Memory (LSTM) model \citep{hochreiter1997long}, wherein the cell state is initialized before the DM's first game (10-round interaction with a bot) to a vector estimated during training, while the hidden state is propagated from game to game.\footnote{This method for sharing information among games outperformed several alternatives we considered.} By managing the cell state in this manner, we model the relationship between successive games of the DM against the same expert.
Second, we train a Transformer model \citep{vaswani2017attention} that takes as input the representation of all rounds up to round $t$.
Third, we use Mamba, which is a modern state-space model \cite{gu2024mamba_COLM}.
Lastly, we implement two strong non-sequential models: an XGBoost classifier \citep{chen2016xgboost}, and a fully connected (FC) neural network.\footnote{Additional experimental details and the hyper-parameter tuning procedures are in Appendix \ref{appendix:HPT}.}

\paragraph{Ablation Analysis}

Experiments with these models aim to shed light on the factors that contribute to the positive impact of the simulation. To this end, we consider several variants of the simulation process of \S
\ref{sec:simulation}.  First, we test the impact of the DM's learning rate parameter ($\eta$) on the prediction performance. Then, we examined the effect of the number of simulated agents on the performance of the models, reasoning that a truly effective simulation is one where more simulated data yields better results, at least up to some threshold. Finally, we consider the relative impact of each component of the simulation.

\subsection{Research Questions}

We consider the following research questions:
\textbf{Q1:} Does incorporating simulation data during model training improve the accuracy of decision prediction in OPE scenarios for different types of prediction models? 
\textbf{Q2:} 
How do the different learning models perform on the human choice prediction task?
\textbf{Q3:} 
Does simulation improve prediction for different types of expert strategies?
\textbf{Q4:} What are the components of the simulation that lead to the improved results?  and 
\textbf{Q5:} How does the representation of the language in the prediction model (EFs vs. plain LLMs) affect the prediction quality with and without simulation?

\section{Results}
\label{chap:results}
\begin{figure*}[h]

  \centering
  \includegraphics[width=\textwidth]{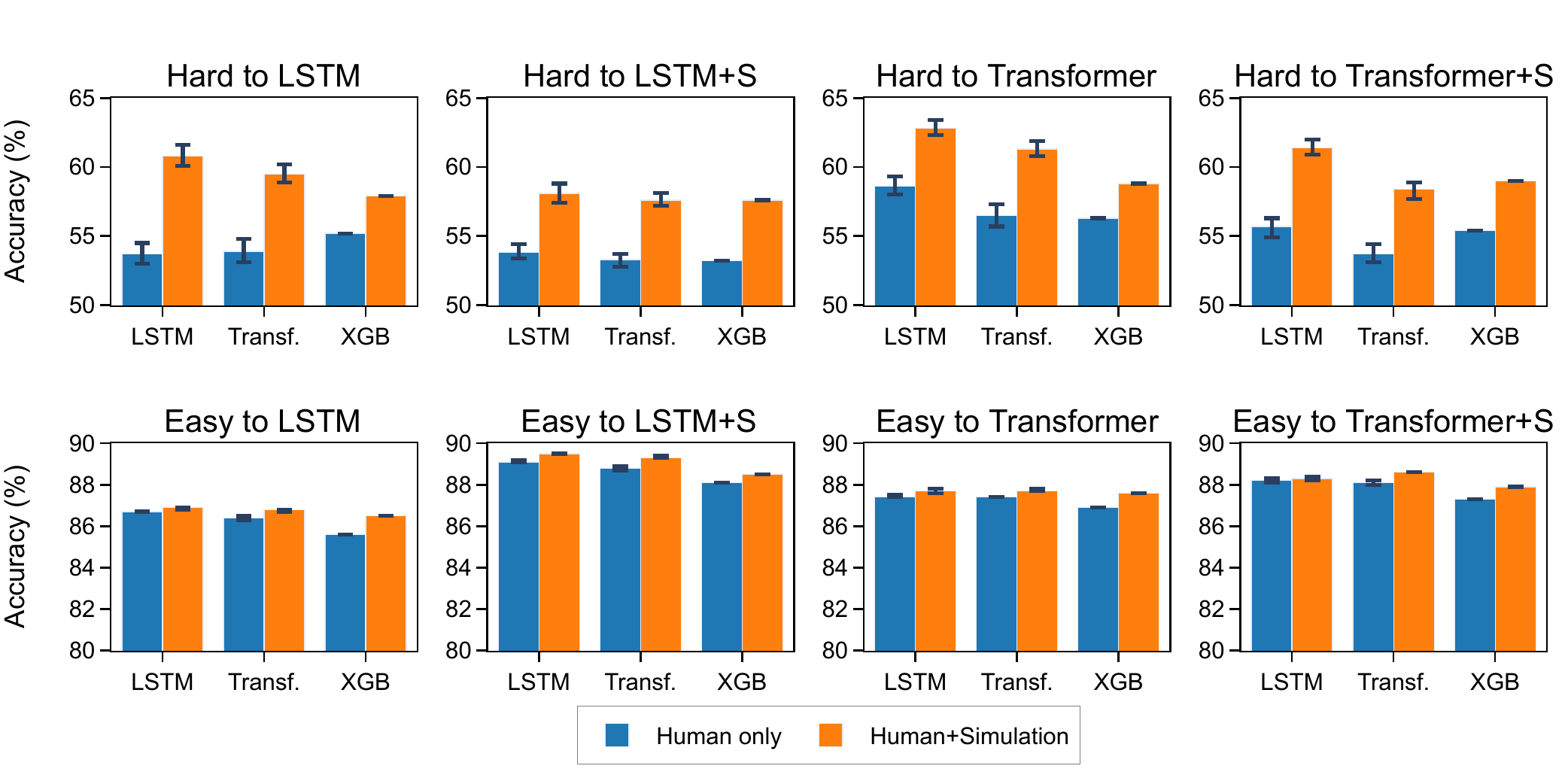}
  \caption{Performance of the models on different sets of hard (top)  and easy (bottom) examples, with 95\% bootstrap confidence intervals. Training on a combination of human-bot interaction and simulated data improves model performance on the hard sets without harming their performance on the easy sets.}
\label{fig:hard-vs-easy}

\end{figure*}


In this section we report the results of each model as the mean of the average accuracy per player (DM) and expert strategy. We average over DMs rather than over decisions so that human DMs who played more games than others are not over-represented in the results.
For the models trained on human-bot interactions only (i.e., without simulation data), LSTM and Mamba show the best performance, outperforming Transformer.
The XGB, FC and Majority models are inferior, and hence in what follows we mostly focus on the results of the LSTM and the Transformer, and the full experimental results can be found in Appendix \ref{app:main-table}.
The rationale behind focusing on LSTM and Transformer (instead of the two best-performing models, LSTM and Mamba) is that the two architectures reflect two extreme modeling approaches (see discussion in Q2), while Mamba conceptually serves as a middle-ground, combining the sequential processing of LSTM with the and scalability of Transformers.

\begin{figure*}[h]
  \centering
  \includegraphics[width=\textwidth]{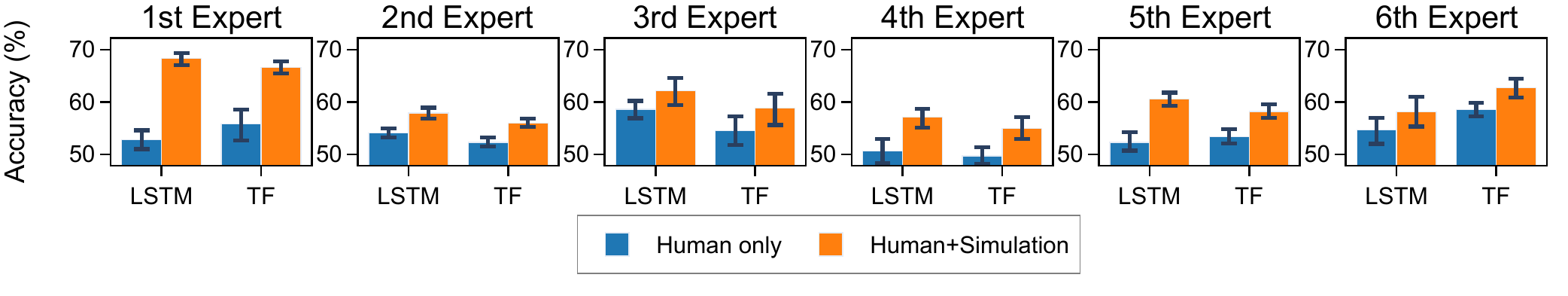}
  \caption{Model performance on the LSTM hard examples, for each expert strategy, with 95\% bootstrap confidence intervals. Training on human and simulation data improves performance for all strategies.}

\label{fig:accuracy-by-sender}
\end{figure*}

\paragraph{The Impact of the Simulation (Q1)}

Figure \ref{fig:hard-vs-easy}, as well as Table \ref{main-table} in appendix \ref{app:main-table}, present the accuracy of each model, when trained on the human interaction data only, and when simulation data is added to the training set (the \emph{+S} models). Our analysis distinguishes \emph{hard} from \emph{easy} examples for each model, and also provides results over the entire test set.
We define a \emph{hard} example for a deep learning model as one for which not all of its 15 variants, differing in their randomly initialized training weights, agree with each other. For the XGB and Majority vote models, we consider examples with confidence levels 40\%-60\% as hard. Non-hard are considered \emph{easy}.\footnote{The results in Figures \ref{fig:hard-vs-easy} and \ref{fig:accuracy-by-sender} (as well as Table \ref{main-table} in appendix \ref{app:main-table}) are presented with 95\% bootstrap confidence intervals, based on the accuracy obtained by all 15 prediction models' variants.} 

For all classifiers and for all hard example sets, combining human and simulated data demonstrates improved performance compared to training on human interaction data only, without harming the prediction on the easy example sets. That is, for each prediction model (column in Figure \ref{fig:hard-vs-easy}), adding simulated data increases accuracy on hard cases (top row) and does not decrease accuracy on easy cases (bottom row).
Specifically, for LSTM the improvement on its own hard example set is 7.1\%, from 53.7\% to 60.8\% accuracy. For the Transformer the corresponding improvement is 4.8\%, from 56.5\% to 61.3\%.

We emphasize that training on the simulated data only yields disappointing performance. For example, on the entire test-set the accuracy of LSTM+S is 83.6\% and of Transformer+S is 83.4\%, and for the LSTM and Transformer models, trained on human-interaction data only, the corresponding numbers are 82.6\% and 82.3\%. At the same time, if we train the LSTM and the Transformer on simulated data only, their accuracy is 78.6\% for LSTM and 78.7\% for the Transformer. Hence, we do not consider simulation-only training any further in this paper.\footnote{In Appendix \ref{app: sim vs human contrib} we compare the contribution of simulated data to the hypothetical case where additional human data is available.}

\paragraph{The Impact of the Prediction Model (Q2)} 
Figure \ref{fig:hard-vs-easy} reveals an insightful observation on the relative effectiveness of different model architectures. Although, as expected, XGB performs worse than both LSTM and Transformer models, an unexpected result is that LSTM often outperforms Transformer in harder cases. Specifically, across the top row of results (excluding 'Hard to LSTM+S'), LSTM consistently outperforms Transformer. This observation holds even in 'Hard to LSTM' cases, where one would anticipate Transformer to perform better, given that these cases are particularly challenging for LSTM.
\begin{figure*}[h]
  \centering
  \includegraphics[width=0.815\textwidth]{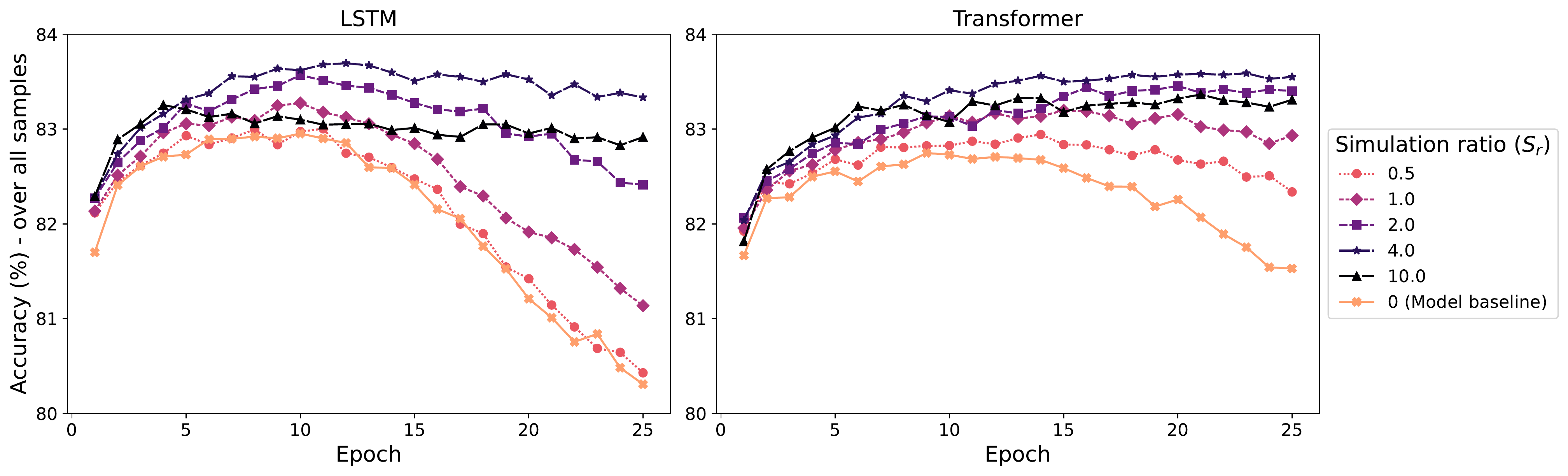}
  \caption{Model performance as a function of the number of epochs, for various values of $S_r$, the ratio between the number of simulated and human DMs in the training set (the number of human interaction examples is fixed to the entire human interaction training set).}
\label{fig:size_effect}
\end{figure*}

A possible explanation lies in the distinct architectural characteristics of LSTM and Transformer. Unlike Transformers, which models dependencies across all elements of an input sequence, LSTMs have an inherent \emph{inductive bias} that encourages reliance on recent sequence elements for prediction. This inductive bias appears advantageous in choice prediction tasks, where human decision-makers (DMs) are generally influenced by recent interactions (a cognitive bias famously known as the "recency effect", see \citealp{ebbinghaus1913contribution}). While it might seem plausible for Transformer to learn such patterns autonomously, it is essential to note that, unlike language modeling tasks, human choice prediction datasets are often relatively small, lacking sufficient data for the Transformer to independently capture such temporal patterns without architectural guidance. Given the constraints on collecting human data—stemming from privacy, budget, and logistical challenges—limited training data is a frequent issue in human choice prediction tasks. This limitation underscores the importance of model selection in addressing these challenges.
\begin{figure*}[h]
  \centering
  \includegraphics[width=0.815\textwidth]{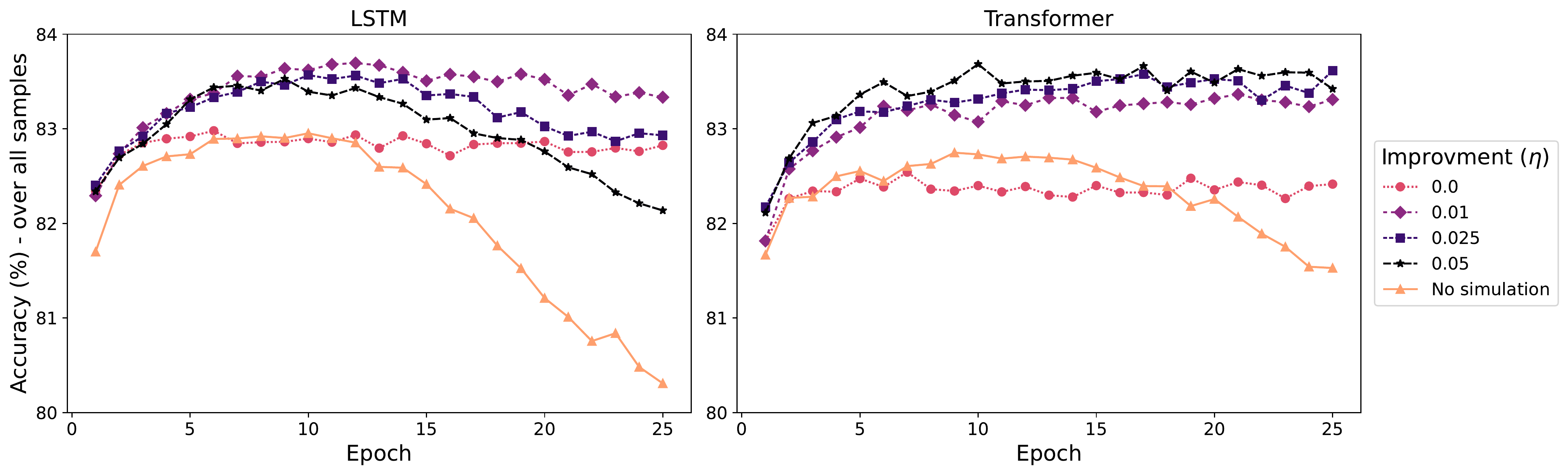}
  \caption{Model performance as a function of the number of epochs, for different values of $\eta$, the improvement rate parameter.}
\label{fig:gamma_effect}
\end{figure*}

\paragraph{The Impact of the Expert Strategy (Q3)}
We highlight that the effectiveness of an expert strategy heavily depends on the behavior of the opponent DM. Different human players respond differently, creating varied effects that impact how accurately their actions can be predicted.
For example, consider a greedy expert who always presents the most positive review. Some DMs may trust and follow consistently positive reviews, while others may learn to disregard them as uninformative. A different expert strategy is likely to dramatically change the behavior of DMs.
Thus, the particular expert strategy (against whom human behavior is predicted) directly influences the complexity and outcome of the prediction task.
Figure \ref{fig:accuracy-by-sender} presents the accuracy of the models, trained with and without simulation, for each of the expert strategies for the set of hard examples of the LSTM model. Apparently, including the simulation data improves the performance of both LSTM and Transformer for each of the expert strategies. The same results are observed when testing the models on the hard Transformer examples. This result increases our confidence in the simulation as it improves performance when considering each strategy separately.

\paragraph{Ablation Analysis (Q4)}

Figure \ref{fig:size_effect} presents model performance as a function of the number of training epochs, for various values of the $S_r$ parameter, which defines the ratio between the number of simulated and human DMs in the training set.\footnote{The number of human interaction examples is fixed to the entire human interaction training set, so higher $S_r$ values simply mean more simulated data.} It can be seen that adding more simulated data improves performance, up to a limit of $S_r = 4$. However, the improvement with $S_r = 10$ is lower than with $S_r = 4$, so the impact of the simulation data is not unlimited. The regularization effect of the simulation is also observed. These results demonstrate the quality of our simulation, as we would expect that more high-quality data would increase its positive impact.

We next examine the relative importance of the various simulation components for improved performance. We start by training the models with different values of $\eta$, the DM improvement rate parameter, which controls the DM learning from experience. Figure \ref{fig:gamma_effect} demonstrates that when $\eta$ = 0, the simulation data neither improves nor harms model performance, and serves only as a means of regularization. For $\eta > 0$, both LSTM and Transformer benefit from the simulation to a similar extent. These results emphasize the importance of learning from experience for OPE, particularly independently of the expert strategy.

Table \ref{tab:sim_type} quantifies the impact of the three heuristics (Language-based, Truthful and Random) on the performance of the LSTM prediction model. The table reveals that all three strategies have a substantial impact on the performance. For example, if we include only one, the accuracy of the eventual prediction model is 70.9\% for language-based, 71.5\% for trustful, and only 10.9\% for random (percentage taken from the prediction accuracy of the complete simulation). The patterns for the Transformer are very similar. We also evaluated the model’s performance when the simulation relies solely on an oracle. In this case, the improvement reaches only 28.8\%.

\begin{table}[h]
\footnotesize
\begin{tabular}{ccc}
\multicolumn{3}{c}{With Random}                                                                                       \\ \cline{2-3} 
\multicolumn{1}{c|}{}                    & \multicolumn{1}{c|}{With Trustful} & \multicolumn{1}{c|}{Without Trustful} \\ \hline
\multicolumn{1}{|c|}{With Lang.-based}    & \multicolumn{1}{c|}{100.0\%}          & \multicolumn{1}{c|}{80.8\%}             \\ \hline
\multicolumn{1}{|c|}{Without Lang.-based} & \multicolumn{1}{c|}{50.3\%}           & \multicolumn{1}{c|}{10.9\%}             \\ \hline
\multicolumn{1}{l}{}                     & \multicolumn{1}{l}{}               & \multicolumn{1}{l}{}                  \\
\multicolumn{3}{c}{Without Random}                                                                                    \\ \cline{2-3} 
\multicolumn{1}{c|}{}                    & \multicolumn{1}{c|}{With Trustful} & \multicolumn{1}{c|}{Without Trustful} \\ \hline
\multicolumn{1}{|c|}{With Lang.-based}    & \multicolumn{1}{c|}{89.2\%}          & \multicolumn{1}{c|}{70.9\%}             \\ \hline
\multicolumn{1}{|c|}{Without Lang.-based} & \multicolumn{1}{c|}{71.5\%}           & \multicolumn{1}{c|}{0\%}  \\ \hline
\end{tabular}
\caption{The impact of simulation heuristics on the LSTM prediction performance (with $\eta = 0.01$ selected via hyperparameter tuning). Percentages are taken from the prediction accuracy of the complete simulation.}
\label{tab:sim_type}
\end{table}

\paragraph{Language Representation Analysis (Q5)}


\begin{figure*}[t]
  \centering
  \includegraphics[width=0.97\textwidth]{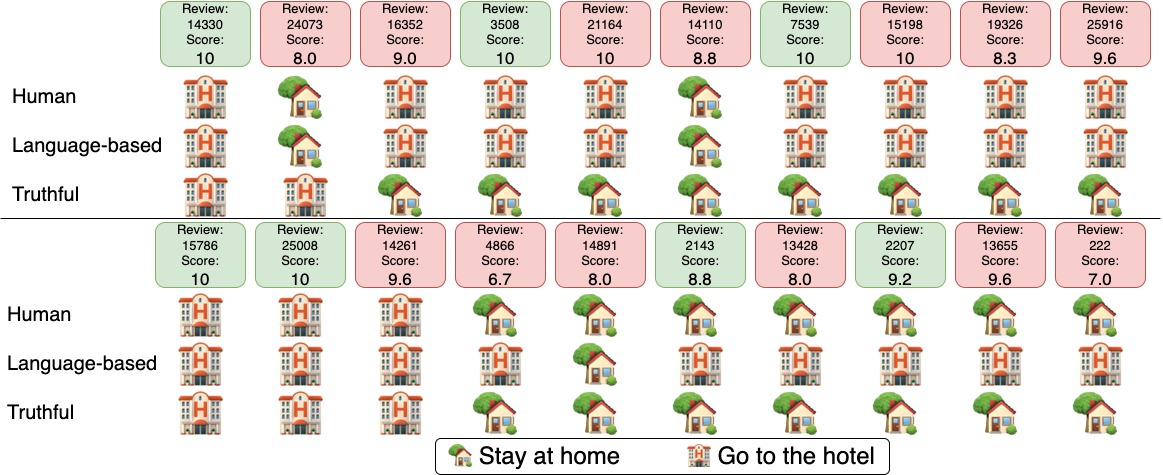}
  \caption{\todo{Two games from the dataset. The top row shows the reviews sent by the agent (score and index) and the hotel quality (green: good, red: bad). The following rows present the human player's decisions and the decisions that would have been made by a simulated DM under the language-based profile and by a simulated DM under the truthful profile in the given situations. In the first game, the human player matched the language-based strategy, while in the second, they followed the truthful strategy.}}
    \label{fig:two_games}
\end{figure*}

Figure \ref{fig:features} (in Appendix \ref{app:lang-representation}) presents the LSTM performance for the three review representation schemes: BERT, GPT-4 and EF, and for various values of the simulation ratio parameter $S_r$.\footnote{For BERT and GPT-4 we take the sentence embeddings and perform dimensionality reduction with PCA to 36 coordinates, to balance the vector size with the number of reviews (3000). 36 is also the number of the EFs.} Evidently, the performance with EF is superior.\footnote{The pattern for the Transformer is similar.}

\section{Human-Simulation Comparison}
\todo{As mentioned in \S \ref{sec:simulation}, we constructed the simulation based on principles that we believe characterize human decision-making. In this section, we examine whether the simulation's behavior aligns with human behavior.}

\subsection{Comparing Simulation Behavior and Human Decision-Making} 
\todo{To demonstrate the similarity between the simulation and human decision-making, we analyzed two vectors: one representing the percentage of times a player chooses to go to a hotel based on a given review, and another capturing the percentage of times a player makes this choice based on the decision history and outcomes from the previous two rounds. Table \ref{tab:similarity_between_simulations_to_human} presents the Pearson correlation coefficients between vectors derived from all human player decisions and a vector constructed from half a million rounds played by simulated decision-makers. We report the correlations for the different heuristics used in the simulation, their combination with the Oracle strategy, and the full simulation, which includes all the strategies.
The results indicate a strong correlation (Pearson coefficient $\ge$ 0.67) between the simulation's decisions and those of human players across both language-based and truthful simulations. In all cases, incorporating an oracle further enhances alignment with human decision-making.
Interestingly, the full simulation, incorporating all strategies, shows lower correlation with humans than the Trustful and Language-based (w. or w/o Oracle) strategies. This contrasts with Table 3, where the full simulation proves superior as training data for prediction. We hypothesize that its random component, while uncorrelated with human behavior, enhances prediction by improving the robustness of the trained predictor.
Figure \ref{fig:two_games} presents two games from the dataset, comparing the human player's behavior to that of Simulated DMs employing language-based and truthful strategies.}

\begin{table}[h]
\begin{tabular}{ccc}
\hline
Simulation Heuristic       & Review  & History  \\ \hline
Oracle                 & 0.52              & 0.23               \\ \hline
Truthful                & 0.69              & 0.67               \\
Truthful + Oracle       & 0.72              & 0.69               \\ \hline
Language-based          & 0.79              & 0.70               \\
Language-based + Oracle & 0.81              & 0.79              \\ \hline
Random                  & -0.00             & 0.06               \\
Random + Oracle         & 0.54              & -0.01              \\ \hline
Full Simulation         & 0.79              & 0.65              \\ \hline
\end{tabular}
\caption{\todo{Pearson correlation coefficients measuring the similarity between human behavior and heuristics in the simulation, based on average decision-making probabilities given a specific review (the Review column) and given the decision and outcome history from the previous two rounds (the History column).}}
\label{tab:similarity_between_simulations_to_human}
\end{table}

\subsection{Improvement Over Time in Human DMs}
\label{app:human learning}
\begin{figure}[h!]
  \centering
  \includegraphics[width=0.88\columnwidth]{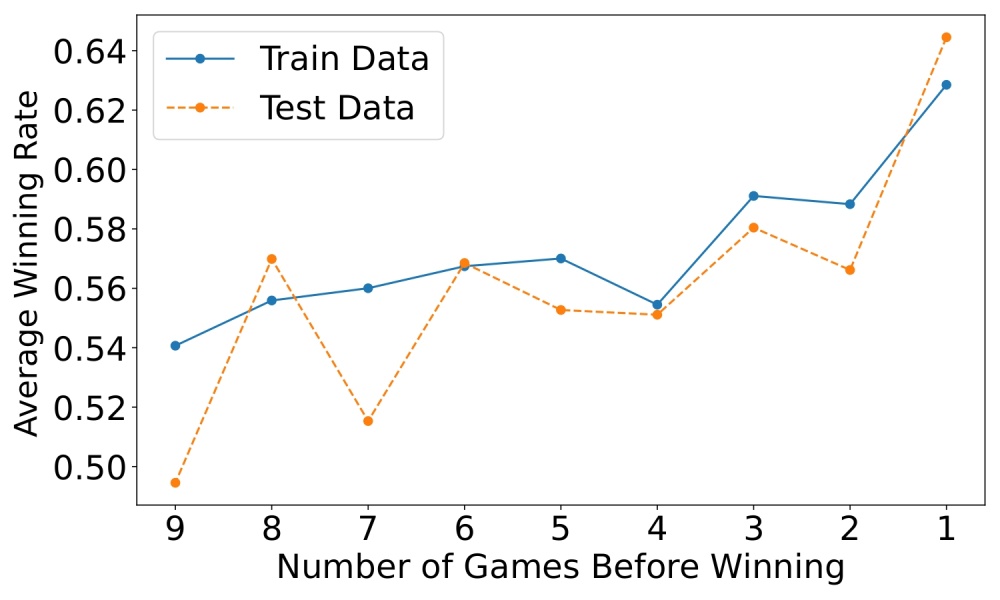}
  \caption{\todo{Round winning probability as a function of the distance from the final game against the same agent.}}
    \label{fig:improvement_over_time}
\end{figure}

\fromApp{
One of the most important assumptions in the simulation is that humans learn over time. In this subsection, we validate this hypothesis. Figure \ref{fig:improvement_over_time} shows the improvement of human players over time, against both training strategies ($E_\mathcal{A}$) and test strategies ($E_\mathcal{B}$). Each point shows the average winning rate (i.e., the probability of taking the "right" action) across all DMs, experts and rounds, in the $i'th$ game before the DM defeats the expert (i.e., reaches the target payoff). The graph clearly shows that as time progresses, DMs are more likely to make correct decisions.}

\section{Discussion}

We addressed the challenge of OPE in language-based persuasion games by proposing a simulation where DMs employ a mixture of interpretable, human-like heuristics that incorporate both behavioral and language-based signals. Additionally, the probability of making correct decisions increases progressively over time. Combining this simulation data with human-bot interactions demonstrated significant improvements.

\paragraph{Limitations and Future Work}
We made several restricting assumptions, which also serve as potential future research directions. First, while our strategy space is rigorously defined and allows us to crystallize the approach, considering more involved expert strategies is a natural extension in bridging this topic into practice. Future work would aim to demonstrate results for a wide range of strategy sets that would serve as $E_\mathcal{A}$ and $E_\mathcal{B}$, and with a larger variety of parameters (e.g., for games with a larger number of rounds). However, since data collection from humans is costly and laborious, we restrict our experiments to a specific choice of $E_\mathcal{A}$ and $E_\mathcal{B}$, potentially affecting the generality of our findings.

Second, extending our approach beyond the persuasion game framework is another appealing future direction. This paper takes an initial step toward a simulation-based framework for off-policy evaluation, demonstrated through a game inspired by economic theory with relevance to domains like recommender systems and e-commerce. Our approach models a decision-making player using basic heuristics—Trustful, Language-based, and Random—while allowing gradual improvement via an “oracle” strategy over time. Although focused on persuasion games, this method could apply broadly to other economic settings where such interpretable “base” strategies are identifiable.

\paragraph{Ethical and Societal Considerations}

Human choice prediction is a field with profound societal implications. Developing technology for predicting human decisions, particularly in economic contexts, holds both promise and risk. On the positive side, such technologies can enhance consumer welfare in recommendation engines. By accurately predicting consumer behavior, system designers can strike a fair balance between the interests of buyers and sellers, optimizing outcomes for both. However, the same technology can be exploited to manipulate consumers into inefficient trades, greedily maximizing profits at their expense.
Beyond commerce, the capabilities of human choice prediction extend to policy-making and public discourse. Persuasion games can model the public's response to various strategies, and while this can inform better policies, it also risks being used to manipulate public opinion, potentially harming societal welfare.

Simulation-based approaches for human choice prediction can also be used in behavioral economics research. These methods can reduce the costs of large-scale human data collection and enhance predictive accuracy. However, simulations can also introduce biases, such as over-reliance on modeled assumptions or misrepresentation of human variability, potentially leading to skewed predictions and misaligned conclusions.

Given these potential benefits and risks, we advocate for the careful and regulated development and application of human choice prediction technologies. Clear guidelines for data collection and use, as well as the development of these tools, are crucial. This can be achieved through ethical codes in academic research, robust guidelines in tech companies, and government regulations. These measures are essential to ensuring that advancements in human choice prediction serve the greater good of society.

\section*{Acknowledgements}

R. Reichart has been partially supported by a VATAT grant on data science. 
The work of M. Tennenholtz and O. Madmon was funded by the European Research Council
(ERC) under the European Union's Horizon 2020 research and innovation program (grant
agreement n° 740435).
The work of E. Shapira is partially supported by a VATAT grant on data science and by the European Research Council
(ERC) under the European Union's Horizon 2020 research and innovation program (grant
agreement n° 740435).

\bibliographystyle{acl_natbib}
\bibliography{bibdb,references}

\begin{thebibliography}{74}
\expandafter\ifx\csname natexlab\endcsname\relax\def\natexlab#1{#1}\fi

\bibitem[{Ai and Weng(2008)}]{ai2008user}
Hua Ai and Fuliang Weng. 2008.
\newblock \href {https://www.aclweb.org/anthology/W08-0126.pdf} {User
  simulation as testing for spoken dialog systems}.
\newblock In \emph{Proceedings of the 9th SIGdial Workshop on Discourse and
  Dialogue}, pages 164--171.

\bibitem[{Akata et~al.(2023)Akata, Schulz, Coda-Forno, Oh, Bethge, and
  Schulz}]{akata2023playing}
Elif Akata, Lion Schulz, Julian Coda-Forno, Seong~Joon Oh, Matthias Bethge, and
  Eric Schulz. 2023.
\newblock Playing repeated games with large language models.
\newblock \emph{arXiv preprint arXiv:2305.16867}.

\bibitem[{Aletras et~al.(2016)Aletras, Tsarapatsanis, Preoţiuc-Pietro, and
  Lampos}]{aletras_predicting_2016}
Nikolaos Aletras, Dimitrios Tsarapatsanis, Daniel Preoţiuc-Pietro, and
  Vasileios Lampos. 2016.
\newblock \href {https://doi.org/10.7717/peerj-cs.93} {Predicting judicial
  decisions of the {European} {Court} of {Human} {Rights}: a {Natural}
  {Language} {Processing} perspective}.
\newblock \emph{PeerJ Computer Science}, 2:e93.
\newblock Publisher: PeerJ Inc.

\bibitem[{Anil et~al.(2023)Anil, Dai, Firat, Johnson, Lepikhin, Passos,
  Shakeri, Taropa, Bailey, Chen, Chu, Clark, Shafey, Huang, Meier-Hellstern,
  Mishra, Moreira, Omernick, Robinson, Ruder, Tay, Xiao, Xu, Zhang, Abrego,
  Ahn, Austin, Barham, Botha, Bradbury, Brahma, Brooks, Catasta, Cheng, Cherry,
  Choquette-Choo, Chowdhery, Crepy, Dave, Dehghani, Dev, Devlin, Díaz, Du,
  Dyer, Feinberg, Feng, Fienber, Freitag, Garcia, Gehrmann, Gonzalez, Gur-Ari,
  Hand, Hashemi, Hou, Howland, Hu, Hui, Hurwitz, Isard, Ittycheriah, Jagielski,
  Jia, Kenealy, Krikun, Kudugunta, Lan, Lee, Lee, Li, Li, Li, Li, Li, Lim, Lin,
  Liu, Liu, Maggioni, Mahendru, Maynez, Misra, Moussalem, Nado, Nham, Ni,
  Nystrom, Parrish, Pellat, Polacek, Polozov, Pope, Qiao, Reif, Richter, Riley,
  Ros, Roy, Saeta, Samuel, Shelby, Slone, Smilkov, So, Sohn, Tokumine, Valter,
  Vasudevan, Vodrahalli, Wang, Wang, Wang, Wang, Wieting, Wu, Xu, Xu, Xue, Yin,
  Yu, Zhang, Zheng, Zheng, Zhou, Zhou, Petrov, and Wu}]{anil_palm_2023}
Rohan Anil, Andrew~M. Dai, Orhan Firat, Melvin Johnson, Dmitry Lepikhin,
  Alexandre Passos, Siamak Shakeri, Emanuel Taropa, Paige Bailey, Zhifeng Chen,
  Eric Chu, Jonathan~H. Clark, Laurent~El Shafey, Yanping Huang, Kathy
  Meier-Hellstern, Gaurav Mishra, Erica Moreira, Mark Omernick, Kevin Robinson,
  Sebastian Ruder, Yi~Tay, Kefan Xiao, Yuanzhong Xu, Yujing Zhang,
  Gustavo~Hernandez Abrego, Junwhan Ahn, Jacob Austin, Paul Barham, Jan Botha,
  James Bradbury, Siddhartha Brahma, Kevin Brooks, Michele Catasta, Yong Cheng,
  Colin Cherry, Christopher~A. Choquette-Choo, Aakanksha Chowdhery, Clément
  Crepy, Shachi Dave, Mostafa Dehghani, Sunipa Dev, Jacob Devlin, Mark Díaz,
  Nan Du, Ethan Dyer, Vlad Feinberg, Fangxiaoyu Feng, Vlad Fienber, Markus
  Freitag, Xavier Garcia, Sebastian Gehrmann, Lucas Gonzalez, Guy Gur-Ari,
  Steven Hand, Hadi Hashemi, Le~Hou, Joshua Howland, Andrea Hu, Jeffrey Hui,
  Jeremy Hurwitz, Michael Isard, Abe Ittycheriah, Matthew Jagielski, Wenhao
  Jia, Kathleen Kenealy, Maxim Krikun, Sneha Kudugunta, Chang Lan, Katherine
  Lee, Benjamin Lee, Eric Li, Music Li, Wei Li, YaGuang Li, Jian Li, Hyeontaek
  Lim, Hanzhao Lin, Zhongtao Liu, Frederick Liu, Marcello Maggioni, Aroma
  Mahendru, Joshua Maynez, Vedant Misra, Maysam Moussalem, Zachary Nado, John
  Nham, Eric Ni, Andrew Nystrom, Alicia Parrish, Marie Pellat, Martin Polacek,
  Alex Polozov, Reiner Pope, Siyuan Qiao, Emily Reif, Bryan Richter, Parker
  Riley, Alex~Castro Ros, Aurko Roy, Brennan Saeta, Rajkumar Samuel, Renee
  Shelby, Ambrose Slone, Daniel Smilkov, David~R. So, Daniel Sohn, Simon
  Tokumine, Dasha Valter, Vijay Vasudevan, Kiran Vodrahalli, Xuezhi Wang,
  Pidong Wang, Zirui Wang, Tao Wang, John Wieting, Yuhuai Wu, Kelvin Xu, Yunhan
  Xu, Linting Xue, Pengcheng Yin, Jiahui Yu, Qiao Zhang, Steven Zheng,
  Ce~Zheng, Weikang Zhou, Denny Zhou, Slav Petrov, and Yonghui Wu. 2023.
\newblock \href {http://arxiv.org/abs/2305.10403} {{PaLM} 2 {Technical}
  {Report}}.
\newblock ArXiv:2305.10403 [cs].

\bibitem[{Apel et~al.(2022)Apel, Erev, Reichart, and
  Tennenholtz}]{apel2020predicting}
Reut Apel, Ido Erev, Roi Reichart, and Moshe Tennenholtz. 2022.
\newblock Predicting decisions in language based persuasion games.
\newblock \emph{Journal of Artificial Intelligence Research}, 73:1025--1091.

\bibitem[{Arora et~al.(2012)Arora, Hazan, and Kale}]{arora2012multiplicative}
Sanjeev Arora, Elad Hazan, and Satyen Kale. 2012.
\newblock The multiplicative weights update method: a meta-algorithm and
  applications.
\newblock \emph{Theory of computing}, 8(1):121--164.

\bibitem[{Auletta et~al.(2023)Auletta, Kallen, di~Bernardo, and
  Richardson}]{auletta2023predicting}
Fabrizia Auletta, Rachel~W Kallen, Mario di~Bernardo, and Michael~J Richardson.
  2023.
\newblock Predicting and understanding human action decisions during skillful
  joint-action using supervised machine learning and explainable-ai.
\newblock \emph{Scientific Reports}, 13(1):4992.

\bibitem[{Aumann et~al.(1995)Aumann, Maschler, and
  Stearns}]{aumann_maschler_stearns_1995}
Robert~John Aumann, Michael~Bahir Maschler, and Richard~E. Stearns. 1995.
\newblock \emph{Repeated games with incomplete information}.
\newblock MIT Press.

\bibitem[{Bahar et~al.(2016)Bahar, Smorodinsky, and
  Tennenholtz}]{bahar2016economic}
Gal Bahar, Rann Smorodinsky, and Moshe Tennenholtz. 2016.
\newblock Economic recommendation systems: One page abstract.
\newblock In \emph{Proceedings of the 2016 ACM Conference on Economics and
  Computation}, pages 757--757.

\bibitem[{Bak and Oh(2018)}]{bak_conversational_2018}
JinYeong Bak and Alice Oh. 2018.
\newblock \href {https://doi.org/10.18653/v1/D18-1115} {Conversational
  {Decision}-{Making} {Model} for {Predicting} the {King}'s {Decision} in the
  {Annals} of the {Joseon} {Dynasty}}.
\newblock In \emph{Proceedings of the 2018 {Conference} on {Empirical}
  {Methods} in {Natural} {Language} {Processing}}, pages 956--961, Brussels,
  Belgium. Association for Computational Linguistics.

\bibitem[{Ben-Porat et~al.(2020)Ben-Porat, Hirsch, Kuchi, Elad, Reichart, and
  Tennenholtz}]{ben2020predicting}
Omer Ben-Porat, Sharon Hirsch, Lital Kuchi, Guy Elad, Roi Reichart, and Moshe
  Tennenholtz. 2020.
\newblock Predicting strategic behavior from free text.
\newblock \emph{Journal of Artificial Intelligence Research}, 68:413--445.

\bibitem[{Bergemann and Morris(2019)}]{BergemannMorris}
Dirk Bergemann and Stephen Morris. 2019.
\newblock Information design: A unified perspective.
\newblock \emph{Journal of Economic Literature}, 57(1):44--95.

\bibitem[{Bourgin et~al.(2019)Bourgin, Peterson, Reichman, Russell, and
  Griffiths}]{bourgin2019cognitive}
David~D Bourgin, Joshua~C Peterson, Daniel Reichman, Stuart~J Russell, and
  Thomas~L Griffiths. 2019.
\newblock Cognitive model priors for predicting human decisions.
\newblock In \emph{International conference on machine learning}, pages
  5133--5141. PMLR.

\bibitem[{Bousmalis et~al.(2018)Bousmalis, Irpan, Wohlhart, Bai, Kelcey,
  Kalakrishnan, Downs, Ibarz, Pastor, Konolige et~al.}]{bousmalis2018using}
Konstantinos Bousmalis, Alex Irpan, Paul Wohlhart, Yunfei Bai, Matthew Kelcey,
  Mrinal Kalakrishnan, Laura Downs, Julian Ibarz, Peter Pastor, Kurt Konolige,
  et~al. 2018.
\newblock Using simulation and domain adaptation to improve efficiency of deep
  robotic grasping.
\newblock In \emph{2018 IEEE international conference on robotics and
  automation (ICRA)}, pages 4243--4250. IEEE.

\bibitem[{Breum et~al.(2024)Breum, Egdal, Mortensen, M{\o}ller, and
  Aiello}]{breum2024persuasive}
Simon~Martin Breum, Daniel~V{\ae}dele Egdal, Victor~Gram Mortensen,
  Anders~Giovanni M{\o}ller, and Luca~Maria Aiello. 2024.
\newblock The persuasive power of large language models.
\newblock In \emph{Proceedings of the International AAAI Conference on Web and
  Social Media}, volume~18, pages 152--163.

\bibitem[{Calderon et~al.(2022)Calderon, Ben-David, Feder, and
  Reichart}]{calderon2022docogen}
Nitay Calderon, Eyal Ben-David, Amir Feder, and Roi Reichart. 2022.
\newblock Docogen: Domain counterfactual generation for low resource domain
  adaptation.
\newblock In \emph{Proceedings of the 60th Annual Meeting of the Association
  for Computational Linguistics (Volume 1: Long Papers)}, pages 7727--7746.

\bibitem[{Cao and Ramezani(2023)}]{cao2022data}
Minh Cao and Ramin Ramezani. 2023.
\newblock Data generation using simulation technology to improve perception
  mechanism of autonomous vehicles.
\newblock In \emph{Journal of Physics: Conference Series}, volume 2547, page
  012006. IOP Publishing.

\bibitem[{Carrasco-Farre(2024)}]{carrascofarre2024largelanguagemodelspersuasive}
Carlos Carrasco-Farre. 2024.
\newblock \href {https://arxiv.org/abs/2404.09329} {Large language models are
  as persuasive as humans, but how? about the cognitive effort and
  moral-emotional language of llm arguments}.
\newblock \emph{arXiv preprint arXiv:2404.09329}.

\bibitem[{Chen and Yang(2021)}]{chen2021weakly}
Jiaao Chen and Diyi Yang. 2021.
\newblock Weakly-supervised hierarchical models for predicting persuasive
  strategies in good-faith textual requests.
\newblock In \emph{Proceedings of the AAAI Conference on Artificial
  Intelligence}, volume~35, pages 12648--12656.

\bibitem[{Chen and Guestrin(2016)}]{chen2016xgboost}
Tianqi Chen and Carlos Guestrin. 2016.
\newblock Xgboost: A scalable tree boosting system.
\newblock In \emph{Proceedings of the 22nd acm sigkdd international conference
  on knowledge discovery and data mining}, pages 785--794.

\bibitem[{Chen et~al.(2023)Chen, Liu, Shan, and Zhong}]{chen2023emergence}
Yiting Chen, Tracy~Xiao Liu, You Shan, and Songfa Zhong. 2023.
\newblock The emergence of economic rationality of gpt.
\newblock \emph{Proceedings of the National Academy of Sciences},
  120(51):e2316205120.

\bibitem[{Chuang et~al.(2023)Chuang, Goyal, Harlalka, Suresh, Hawkins, Yang,
  Shah, Hu, and Rogers}]{chuang2023simulating}
Yun-Shiuan Chuang, Agam Goyal, Nikunj Harlalka, Siddharth Suresh, Robert
  Hawkins, Sijia Yang, Dhavan Shah, Junjie Hu, and Timothy~T Rogers. 2023.
\newblock Simulating opinion dynamics with networks of llm-based agents.
\newblock \emph{arXiv preprint arXiv:2311.09618}.

\bibitem[{Coulom(2006)}]{coulom2006efficient}
R{\'e}mi Coulom. 2006.
\newblock \href {https://link.springer.com/chapter/10.1007/978-3-540-75538-8_7}
  {Efficient selectivity and backup operators in monte-carlo tree search}.
\newblock In \emph{International conference on computers and games}, pages
  72--83. Springer.

\bibitem[{Ebbinghaus(1913)}]{ebbinghaus1913contribution}
Hermann Ebbinghaus. 1913.
\newblock A contribution to experimental psychology.
\newblock \emph{New York, NY: Teachers College, Columbia University}.

\bibitem[{Emek et~al.(2014)Emek, Feldman, Gamzu, PaesLeme, and
  Tennenholtz}]{emek2014signaling}
Yuval Emek, Michal Feldman, Iftah Gamzu, Renato PaesLeme, and Moshe
  Tennenholtz. 2014.
\newblock Signaling schemes for revenue maximization.
\newblock \emph{ACM Transactions on Economics and Computation (TEAC)},
  2(2):1--19.

\bibitem[{Freund and Schapire(1999)}]{freund1999adaptive}
Yoav Freund and Robert~E Schapire. 1999.
\newblock Adaptive game playing using multiplicative weights.
\newblock \emph{Games and Economic Behavior}, 29(1-2):79--103.

\bibitem[{Fudenberg and Levine(1995)}]{fudenberg1995consistency}
Drew Fudenberg and David~K Levine. 1995.
\newblock Consistency and cautious fictitious play.
\newblock \emph{Journal of Economic Dynamics and Control}, 19(5-7):1065--1089.

\bibitem[{Fudenberg and Tirole(1991)}]{fudenberg1991game}
Drew Fudenberg and Jean Tirole. 1991.
\newblock \href {https://mitpress.mit.edu/books/game-theory} {Game theory}.
\newblock \emph{Cambridge, Massachusetts}, 393(12):80.

\bibitem[{Gigerenzer and Brighton(2009)}]{gigerenzer2009homo}
Gerd Gigerenzer and Henry Brighton. 2009.
\newblock Homo heuristicus: Why biased minds make better inferences.
\newblock \emph{Topics in cognitive science}, 1(1):107--143.

\bibitem[{Gonz{\'a}lez et~al.(2010)Gonz{\'a}lez, Quarteroni, Riccardi, and
  Varges}]{gonzalez2010cooperative}
Meritxell Gonz{\'a}lez, Silvia Quarteroni, Giuseppe Riccardi, and Sebastian
  Varges. 2010.
\newblock \href {https://www.aclweb.org/anthology/W10-4338.pdf} {Cooperative
  user models in statistical dialog simulators}.
\newblock In \emph{Proceedings of the SIGDIAL 2010 Conference}, pages 217--220.

\bibitem[{Gu and Dao(2024)}]{gu2024mamba_COLM}
Albert Gu and Tri Dao. 2024.
\newblock \href {https://openreview.net/forum?id=tEYskw1VY2} {Mamba:
  Linear-time sequence modeling with selective state spaces}.
\newblock In \emph{First Conference on Language Modeling}.

\bibitem[{Guo et~al.(2024)Guo, Bu, Wang, Ren, Sui, Shang, and
  Lu}]{guo2024economics}
Shangmin Guo, Haoran Bu, Haochuan Wang, Yi~Ren, Dianbo Sui, Yuming Shang, and
  Siting Lu. 2024.
\newblock Economics arena for large language models.
\newblock \emph{arXiv preprint arXiv:2401.01735}.

\bibitem[{Hidey and McKeown(2018)}]{hidey2018persuasive}
Christopher Hidey and Kathleen McKeown. 2018.
\newblock \href {https://ojs.aaai.org/index.php/AAAI/article/view/12003}
  {Persuasive influence detection: The role of argument sequencing}.
\newblock In \emph{Proceedings of the AAAI Conference on Artificial
  Intelligence}, volume~32.

\bibitem[{Hidey et~al.(2017)Hidey, Musi, Hwang, Muresan, and
  McKeown}]{hidey_analyzing_2017}
Christopher Hidey, Elena Musi, Alyssa Hwang, Smaranda Muresan, and Kathy
  McKeown. 2017.
\newblock \href {https://doi.org/10.18653/v1/W17-5102} {Analyzing the
  {Semantic} {Types} of {Claims} and {Premises} in an {Online} {Persuasive}
  {Forum}}.
\newblock In \emph{Proceedings of the 4th {Workshop} on {Argument} {Mining}},
  pages 11--21, Copenhagen, Denmark. Association for Computational Linguistics.

\bibitem[{Hiraoka et~al.(2014)Hiraoka, Neubig, Sakti, Toda, and
  Nakamura}]{hiraoka2014reinforcement}
Takuya Hiraoka, Graham Neubig, Sakriani Sakti, Tomoki Toda, and Satoshi
  Nakamura. 2014.
\newblock \href {https://www.aclweb.org/anthology/C14-1161.pdf} {Reinforcement
  learning of cooperative persuasive dialogue policies using framing}.
\newblock In \emph{Proceedings of COLING 2014, the 25th International
  Conference on Computational Linguistics: Technical Papers}, pages 1706--1717.

\bibitem[{Hochreiter and Schmidhuber(1997)}]{hochreiter1997long}
Sepp Hochreiter and J{\"u}rgen Schmidhuber. 1997.
\newblock \href
  {http://citeseerx.ist.psu.edu/viewdoc/download?doi=10.1.1.676.4320&rep=rep1&type=pdf}
  {Long short-term memory}.
\newblock \emph{Neural computation}, 9(8):1735--1780.

\bibitem[{Horton(2023)}]{horton2023large}
John~J. Horton. 2023.
\newblock Large language models as simulated economic agents: What can we learn
  from homo silicus?
\newblock Technical report, National Bureau of Economic Research.

\bibitem[{Hussain et~al.()Hussain, Binz, Mata, and Wulff}]{hussain2023tutorial}
Zak Hussain, Marcel Binz, Rui Mata, and Dirk~U Wulff.
\newblock A tutorial on open-source large language models for behavioral
  science.

\bibitem[{Hutchinson and Gigerenzer(2005)}]{hutchinson2005simple}
John~MC Hutchinson and Gerd Gigerenzer. 2005.
\newblock Simple heuristics and rules of thumb: Where psychologists and
  behavioural biologists might meet.
\newblock \emph{Behavioural processes}, 69(2):97--124.

\bibitem[{Jung et~al.(2008)Jung, Lee, Kim, and Lee}]{jung_integrated_2008}
Sangkeun Jung, Cheongjae Lee, Kyungduk Kim, and Gary~Geunbae Lee. 2008.
\newblock \href {https://aclanthology.org/W08-1503} {An {Integrated} {Dialog}
  {Simulation} {Technique} for {Evaluating} {Spoken} {Dialog} {Systems}}.
\newblock In \emph{Coling 2008: {Proceedings} of the workshop on {Speech}
  {Processing} for {Safety} {Critical} {Translation} and {Pervasive}
  {Applications}}, pages 9--16, Manchester, UK. Coling 2008 Organizing
  Committee.

\bibitem[{Kamenica and Gentzkow(2011)}]{kamenica_gentzkow_2009}
Emir Kamenica and Matthew Gentzkow. 2011.
\newblock \href {https://doi.org/10.3386/w15540} {Bayesian persuasion}.
\newblock \emph{American Economic Review}, 101:2590--2615.

\bibitem[{Kenton and Toutanova(2019)}]{kenton2019bert}
Jacob Devlin Ming-Wei~Chang Kenton and Lee~Kristina Toutanova. 2019.
\newblock Bert: Pre-training of deep bidirectional transformers for language
  understanding.
\newblock In \emph{Proceedings of NAACL-HLT}, pages 4171--4186.

\bibitem[{Liu et~al.(2023)Liu, Yang, Jia, Zhang, Zhou, Dai, Yang, and
  Vosoughi}]{liu2023training}
Ruibo Liu, Ruixin Yang, Chenyan Jia, Ge~Zhang, Denny Zhou, Andrew~M Dai, Diyi
  Yang, and Soroush Vosoughi. 2023.
\newblock Training socially aligned language models in simulated human society.
\newblock \emph{arXiv preprint arXiv:2305.16960}.

\bibitem[{Mas-Colell et~al.(1995)Mas-Colell, Whinston, and
  Green}]{mas-colell.whinston.ea95}
Andreu Mas-Colell, Michael~D. Whinston, and Jerry~R. Green. 1995.
\newblock \emph{Microeconomic Theory}.
\newblock Oxford University Press.

\bibitem[{Matz et~al.(2024)Matz, Teeny, Vaid, Peters, Harari, and
  Cerf}]{matz2024potential}
SC~Matz, JD~Teeny, Sumer~S Vaid, H~Peters, GM~Harari, and M~Cerf. 2024.
\newblock The potential of generative ai for personalized persuasion at scale.
\newblock \emph{Scientific Reports}, 14(1):4692.

\bibitem[{Medvedeva et~al.(2020)Medvedeva, Vols, and
  Wieling}]{medvedeva2020using}
Masha Medvedeva, Michel Vols, and Martijn Wieling. 2020.
\newblock \href {https://link.springer.com/article/10.1007/s10506-019-09255-y}
  {Using machine learning to predict decisions of the european court of human
  rights}.
\newblock \emph{Artificial Intelligence and Law}, 28(2):237--266.

\bibitem[{{Meta} et~al.(2022){Meta}, Bakhtin, Brown, Dinan, Farina, Flaherty,
  Fried, Goff, Gray, Hu, Jacob, Komeili, Konath, Kwon, Lerer, Lewis, Miller,
  Mitts, Renduchintala, Roller, Rowe, Shi, Spisak, Wei, Wu, Zhang, and
  Zijlstra}]{meta_human-level_2022}
{Meta}, Anton Bakhtin, Noam Brown, Emily Dinan, Gabriele Farina, Colin
  Flaherty, Daniel Fried, Andrew Goff, Jonathan Gray, Hengyuan Hu, Athul~Paul
  Jacob, Mojtaba Komeili, Karthik Konath, Minae Kwon, Adam Lerer, Mike Lewis,
  Alexander~H. Miller, Sasha Mitts, Adithya Renduchintala, Stephen Roller, Dirk
  Rowe, Weiyan Shi, Joe Spisak, Alexander Wei, David Wu, Hugh Zhang, and Markus
  Zijlstra. 2022.
\newblock \href {https://doi.org/10.1126/science.ade9097} {Human-level play in
  the game of {Diplomacy} by combining language models with strategic
  reasoning}.
\newblock \emph{Science}, 378(6624):1067--1074.
\newblock Publisher: American Association for the Advancement of Science.

\bibitem[{OpenAI et~al.(2023)OpenAI, Achiam, Adler, Agarwal, Ahmad, Akkaya,
  Aleman, Almeida, Altenschmidt, Altman, Anadkat, Avila, Babuschkin, Balaji,
  Balcom, Baltescu, Bao, Bavarian, Belgum, Bello, Berdine, Bernadett-Shapiro,
  Berner, Bogdonoff, Boiko, Boyd, Brakman, Brockman, Brooks, Brundage, Button,
  Cai, Campbell, Cann, Carey, Carlson, Carmichael, Chan, Chang, Chantzis, Chen,
  Chen, Chen, Chen, Chen, Chess, Cho, Chu, Chung, Cummings, Currier, Dai,
  Decareaux, Degry, Deutsch, Deville, Dhar, Dohan, Dowling, Dunning, Ecoffet,
  Eleti, Eloundou, Farhi, Fedus, Felix, Fishman, Forte, Fulford, Gao, Georges,
  Gibson, Goel, Gogineni, Goh, Gontijo-Lopes, Gordon, Grafstein, Gray, Greene,
  Gross, Gu, Guo, Hallacy, Han, Harris, He, Heaton, Heidecke, Hesse, Hickey,
  Hickey, Hoeschele, Houghton, Hsu, Hu, Hu, Huizinga, Jain, Jain, Jang, Jiang,
  Jiang, Jin, Jin, Jomoto, Jonn, Jun, Kaftan, Kaiser, Kamali, Kanitscheider,
  Keskar, Khan, Kilpatrick, Kim, Kim, Kim, Kirchner, Kiros, Knight, Kokotajlo,
  Kondraciuk, Kondrich, Konstantinidis, Kosic, Krueger, Kuo, Lampe, Lan, Lee,
  Leike, Leung, Levy, Li, Lim, Lin, Lin, Litwin, Lopez, Lowe, Lue, Makanju,
  Malfacini, Manning, Markov, Markovski, Martin, Mayer, Mayne, McGrew,
  McKinney, McLeavey, McMillan, McNeil, Medina, Mehta, Menick, Metz,
  Mishchenko, Mishkin, Monaco, Morikawa, Mossing, Mu, Murati, Murk, Mély,
  Nair, Nakano, Nayak, Neelakantan, Ngo, Noh, Ouyang, O'Keefe, Pachocki, Paino,
  Palermo, Pantuliano, Parascandolo, Parish, Parparita, Passos, Pavlov, Peng,
  Perelman, Peres, Petrov, Pinto, Michael, Pokorny, Pokrass, Pong, Powell,
  Power, Power, Proehl, Puri, Radford, Rae, Ramesh, Raymond, Real, Rimbach,
  Ross, Rotsted, Roussez, Ryder, Saltarelli, Sanders, Santurkar, Sastry,
  Schmidt, Schnurr, Schulman, Selsam, Sheppard, Sherbakov, Shieh, Shoker,
  Shyam, Sidor, Sigler, Simens, Sitkin, Slama, Sohl, Sokolowsky, Song,
  Staudacher, Such, Summers, Sutskever, Tang, Tezak, Thompson, Tillet,
  Tootoonchian, Tseng, Tuggle, Turley, Tworek, Uribe, Vallone, Vijayvergiya,
  Voss, Wainwright, Wang, Wang, Wang, Ward, Wei, Weinmann, Welihinda, Welinder,
  Weng, Weng, Wiethoff, Willner, Winter, Wolrich, Wong, Workman, Wu, Wu, Wu,
  Xiao, Xu, Yoo, Yu, Yuan, Zaremba, Zellers, Zhang, Zhang, Zhao, Zheng, Zhuang,
  Zhuk, and Zoph}]{openai_gpt-4_2023}
OpenAI, Josh Achiam, Steven Adler, Sandhini Agarwal, Lama Ahmad, Ilge Akkaya,
  Florencia~Leoni Aleman, Diogo Almeida, Janko Altenschmidt, Sam Altman,
  Shyamal Anadkat, Red Avila, Igor Babuschkin, Suchir Balaji, Valerie Balcom,
  Paul Baltescu, Haiming Bao, Mo~Bavarian, Jeff Belgum, Irwan Bello, Jake
  Berdine, Gabriel Bernadett-Shapiro, Christopher Berner, Lenny Bogdonoff, Oleg
  Boiko, Madelaine Boyd, Anna-Luisa Brakman, Greg Brockman, Tim Brooks, Miles
  Brundage, Kevin Button, Trevor Cai, Rosie Campbell, Andrew Cann, Brittany
  Carey, Chelsea Carlson, Rory Carmichael, Brooke Chan, Che Chang, Fotis
  Chantzis, Derek Chen, Sully Chen, Ruby Chen, Jason Chen, Mark Chen, Ben
  Chess, Chester Cho, Casey Chu, Hyung~Won Chung, Dave Cummings, Jeremiah
  Currier, Yunxing Dai, Cory Decareaux, Thomas Degry, Noah Deutsch, Damien
  Deville, Arka Dhar, David Dohan, Steve Dowling, Sheila Dunning, Adrien
  Ecoffet, Atty Eleti, Tyna Eloundou, David Farhi, Liam Fedus, Niko Felix,
  Simón~Posada Fishman, Juston Forte, Isabella Fulford, Leo Gao, Elie Georges,
  Christian Gibson, Vik Goel, Tarun Gogineni, Gabriel Goh, Rapha Gontijo-Lopes,
  Jonathan Gordon, Morgan Grafstein, Scott Gray, Ryan Greene, Joshua Gross,
  Shixiang~Shane Gu, Yufei Guo, Chris Hallacy, Jesse Han, Jeff Harris, Yuchen
  He, Mike Heaton, Johannes Heidecke, Chris Hesse, Alan Hickey, Wade Hickey,
  Peter Hoeschele, Brandon Houghton, Kenny Hsu, Shengli Hu, Xin Hu, Joost
  Huizinga, Shantanu Jain, Shawn Jain, Joanne Jang, Angela Jiang, Roger Jiang,
  Haozhun Jin, Denny Jin, Shino Jomoto, Billie Jonn, Heewoo Jun, Tomer Kaftan,
  Łukasz Kaiser, Ali Kamali, Ingmar Kanitscheider, Nitish~Shirish Keskar,
  Tabarak Khan, Logan Kilpatrick, Jong~Wook Kim, Christina Kim, Yongjik Kim,
  Hendrik Kirchner, Jamie Kiros, Matt Knight, Daniel Kokotajlo, Łukasz
  Kondraciuk, Andrew Kondrich, Aris Konstantinidis, Kyle Kosic, Gretchen
  Krueger, Vishal Kuo, Michael Lampe, Ikai Lan, Teddy Lee, Jan Leike, Jade
  Leung, Daniel Levy, Chak~Ming Li, Rachel Lim, Molly Lin, Stephanie Lin,
  Mateusz Litwin, Theresa Lopez, Ryan Lowe, Patricia Lue, Anna Makanju, Kim
  Malfacini, Sam Manning, Todor Markov, Yaniv Markovski, Bianca Martin, Katie
  Mayer, Andrew Mayne, Bob McGrew, Scott~Mayer McKinney, Christine McLeavey,
  Paul McMillan, Jake McNeil, David Medina, Aalok Mehta, Jacob Menick, Luke
  Metz, Andrey Mishchenko, Pamela Mishkin, Vinnie Monaco, Evan Morikawa, Daniel
  Mossing, Tong Mu, Mira Murati, Oleg Murk, David Mély, Ashvin Nair, Reiichiro
  Nakano, Rajeev Nayak, Arvind Neelakantan, Richard Ngo, Hyeonwoo Noh, Long
  Ouyang, Cullen O'Keefe, Jakub Pachocki, Alex Paino, Joe Palermo, Ashley
  Pantuliano, Giambattista Parascandolo, Joel Parish, Emy Parparita, Alex
  Passos, Mikhail Pavlov, Andrew Peng, Adam Perelman, Filipe de Avila~Belbute
  Peres, Michael Petrov, Henrique Ponde de~Oliveira Pinto, Michael, Pokorny,
  Michelle Pokrass, Vitchyr Pong, Tolly Powell, Alethea Power, Boris Power,
  Elizabeth Proehl, Raul Puri, Alec Radford, Jack Rae, Aditya Ramesh, Cameron
  Raymond, Francis Real, Kendra Rimbach, Carl Ross, Bob Rotsted, Henri Roussez,
  Nick Ryder, Mario Saltarelli, Ted Sanders, Shibani Santurkar, Girish Sastry,
  Heather Schmidt, David Schnurr, John Schulman, Daniel Selsam, Kyla Sheppard,
  Toki Sherbakov, Jessica Shieh, Sarah Shoker, Pranav Shyam, Szymon Sidor, Eric
  Sigler, Maddie Simens, Jordan Sitkin, Katarina Slama, Ian Sohl, Benjamin
  Sokolowsky, Yang Song, Natalie Staudacher, Felipe~Petroski Such, Natalie
  Summers, Ilya Sutskever, Jie Tang, Nikolas Tezak, Madeleine Thompson, Phil
  Tillet, Amin Tootoonchian, Elizabeth Tseng, Preston Tuggle, Nick Turley,
  Jerry Tworek, Juan Felipe~Cerón Uribe, Andrea Vallone, Arun Vijayvergiya,
  Chelsea Voss, Carroll Wainwright, Justin~Jay Wang, Alvin Wang, Ben Wang,
  Jonathan Ward, Jason Wei, C.~J. Weinmann, Akila Welihinda, Peter Welinder,
  Jiayi Weng, Lilian Weng, Matt Wiethoff, Dave Willner, Clemens Winter, Samuel
  Wolrich, Hannah Wong, Lauren Workman, Sherwin Wu, Jeff Wu, Michael Wu, Kai
  Xiao, Tao Xu, Sarah Yoo, Kevin Yu, Qiming Yuan, Wojciech Zaremba, Rowan
  Zellers, Chong Zhang, Marvin Zhang, Shengjia Zhao, Tianhao Zheng, Juntang
  Zhuang, William Zhuk, and Barret Zoph. 2023.
\newblock \href {https://doi.org/10.48550/arXiv.2303.08774} {{GPT}-4
  {Technical} {Report}}.
\newblock ArXiv:2303.08774 [cs].

\bibitem[{Oroojlooy and Hajinezhad(2022)}]{oroojlooy2022review}
Afshin Oroojlooy and Davood Hajinezhad. 2022.
\newblock A review of cooperative multi-agent deep reinforcement learning.
\newblock \emph{Applied Intelligence}, pages 1--46.

\bibitem[{Oved et~al.(2020)Oved, Feder, and Reichart}]{nadavAmirRoi2020}
Nadav Oved, Amir Feder, and Roi Reichart. 2020.
\newblock \href {https://doi.org/10.1162/coli\_a\_00383} {Predicting in-game
  actions from interviews of nba players}.
\newblock \emph{Computational Linguistics}, 46(3):667--712.

\bibitem[{Park et~al.(2023)Park, O'Brien, Cai, Ringel~Morris, Liang, and
  Bernstein}]{Park:23}
Joon~Sung Park, Joseph~C. O'Brien, Carrie~J. Cai, Meredith Ringel~Morris, Percy
  Liang, and Michael~S. Bernstein. 2023.
\newblock Generative agents: Interactive simulacra of human behavior.
\newblock In \emph{The 36th Annual ACM Symposium on User Interface Software and
  Technology}, pages 1--22.

\bibitem[{Plonsky et~al.(2019)Plonsky, Apel, Ert, Tennenholtz, Bourgin,
  Peterson, Reichman, Griffiths, Russell, Carter
  et~al.}]{plonsky2019predicting}
Ori Plonsky, Reut Apel, Eyal Ert, Moshe Tennenholtz, David Bourgin, Joshua~C
  Peterson, Daniel Reichman, Thomas~L Griffiths, Stuart~J Russell, Evan~C
  Carter, et~al. 2019.
\newblock Predicting human decisions with behavioral theories and machine
  learning.
\newblock \emph{arXiv preprint arXiv:1904.06866}.

\bibitem[{Plonsky et~al.(2017)Plonsky, Erev, Hazan, and
  Tennenholtz}]{psyforest_plonsky_2017}
Ori Plonsky, Ido Erev, Tamir Hazan, and Moshe Tennenholtz. 2017.
\newblock \href {http://aaai.org/ocs/index.php/AAAI/AAAI17/paper/view/14925}
  {Psychological forest: Predicting human behavior}.
\newblock In \emph{Proceedings of the Thirty-First {AAAI} Conference on
  Artificial Intelligence, February 4-9, 2017, San Francisco, California,
  {USA}}, pages 656--662. {AAAI} Press.

\bibitem[{Raifer et~al.(2022)Raifer, Rotman, Apel, Tennenholtz, and
  Reichart}]{raifer2022designing}
Maya Raifer, Guy Rotman, Reut Apel, Moshe Tennenholtz, and Roi Reichart. 2022.
\newblock Designing an automatic agent for repeated language--based persuasion
  games.
\newblock \emph{Transactions of the Association for Computational Linguistics},
  10:307--324.

\bibitem[{Rosenfeld and Kraus(2018)}]{rosenfeld2018predicting}
Ariel Rosenfeld and Sarit Kraus. 2018.
\newblock Predicting human decision-making: From prediction to action.
\newblock \emph{Synthesis lectures on artificial intelligence and machine
  learning}, 12(1):1--150.

\bibitem[{Saadallah et~al.(2022)Saadallah, Finkeldey, Bu{\ss}, Morik,
  Wiederkehr, and Rhode}]{saadallah2022simulation}
Amal Saadallah, Felix Finkeldey, Jens Bu{\ss}, Katharina Morik, Petra
  Wiederkehr, and Wolfgang Rhode. 2022.
\newblock Simulation and sensor data fusion for machine learning application.
\newblock \emph{Advanced Engineering Informatics}, 52:101600.

\bibitem[{Schrittwieser et~al.(2020)Schrittwieser, Antonoglou, Hubert,
  Simonyan, Sifre, Schmitt, Guez, Lockhart, Hassabis, Graepel, Lillicrap, and
  Silver}]{SchrittwieserAH20}
Julian Schrittwieser, Ioannis Antonoglou, Thomas Hubert, Karen Simonyan,
  Laurent Sifre, Simon Schmitt, Arthur Guez, Edward Lockhart, Demis Hassabis,
  Thore Graepel, Timothy~P. Lillicrap, and David Silver. 2020.
\newblock Mastering atari, go, chess and shogi by planning with a learned
  model.
\newblock \emph{Nat.}, 588(7839):604--609.

\bibitem[{Shapira et~al.(2024{\natexlab{a}})Shapira, Madmon, Reichart, and
  Tennenholtz}]{shapira_can_2024}
Eilam Shapira, Omer Madmon, Roi Reichart, and Moshe Tennenholtz.
  2024{\natexlab{a}}.
\newblock \href {https://doi.org/10.48550/arXiv.2401.17435} {Can {Large}
  {Language} {Models} {Replace} {Economic} {Choice} {Prediction} {Labs}?}
\newblock ArXiv:2401.17435 [cs].

\bibitem[{Shapira et~al.(2024{\natexlab{b}})Shapira, Madmon, Reinman, Amouyal,
  Reichart, and Tennenholtz}]{shapira2024glee}
Eilam Shapira, Omer Madmon, Itamar Reinman, Samuel~Joseph Amouyal, Roi
  Reichart, and Moshe Tennenholtz. 2024{\natexlab{b}}.
\newblock Glee: A unified framework and benchmark for language-based economic
  environments.
\newblock \emph{arXiv preprint arXiv:2410.05254}.

\bibitem[{Shi et~al.(2019)Shi, Qian, Wang, and Yu}]{shi_how_2019}
Weiyan Shi, Kun Qian, Xuewei Wang, and Zhou Yu. 2019.
\newblock \href {https://doi.org/10.18653/v1/D19-1206} {How to {Build} {User}
  {Simulators} to {Train} {RL}-based {Dialog} {Systems}}.
\newblock In \emph{Proceedings of the 2019 {Conference} on {Empirical}
  {Methods} in {Natural} {Language} {Processing} and the 9th {International}
  {Joint} {Conference} on {Natural} {Language} {Processing}
  ({EMNLP}-{IJCNLP})}, pages 1990--2000, Hong Kong, China. Association for
  Computational Linguistics.

\bibitem[{Silver et~al.(2018)Silver, Hubert, Schrittwieser, Antonoglou, Lai,
  Guez, Lanctot, Sifre, Kumaran, Graepel et~al.}]{silveretal}
David Silver, Thomas Hubert, Julian Schrittwieser, Ioannis Antonoglou, Matthew
  Lai, Arthur Guez, Marc Lanctot, Laurent Sifre, Dharshan Kumaran, Thore
  Graepel, et~al. 2018.
\newblock \href {https://science.sciencemag.org/content/362/6419/1140/} {A
  general reinforcement learning algorithm that masters chess, shogi, and go
  through self-play}.
\newblock \emph{Science}, 362(6419):1140--1144.

\bibitem[{Silver et~al.(2017)Silver, Schrittwieser, Simonyan, Antonoglou,
  Huang, Guez, Hubert, Baker, Lai, Bolton, Chen, Lillicrap, Hui, Sifre, van~den
  Driessche, Graepel, and Hassabis}]{SilverSSAHGHBLB17}
David Silver, Julian Schrittwieser, Karen Simonyan, Ioannis Antonoglou, Aja
  Huang, Arthur Guez, Thomas Hubert, Lucas Baker, Matthew Lai, Adrian Bolton,
  Yutian Chen, Timothy~P. Lillicrap, Fan Hui, Laurent Sifre, George van~den
  Driessche, Thore Graepel, and Demis Hassabis. 2017.
\newblock Mastering the game of go without human knowledge.
\newblock \emph{Nat.}, 550(7676):354--359.

\bibitem[{Tan et~al.(2016)Tan, Niculae, Danescu-Niculescu-Mizil, and
  Lee}]{tan2016winning}
Chenhao Tan, Vlad Niculae, Cristian Danescu-Niculescu-Mizil, and Lillian Lee.
  2016.
\newblock \href
  {https://dl.acm.org/doi/pdf/10.1145/2872427.2883081?casa_token=xA0jm_smMdEAAAAA:W5I1QG6UMosDHXmwyFtGMa6nWmbYOChX_8FiGhj6sfbJqdCTXmEgkbuEqCT9m6ybeCAIFQ4vFbC1yQ}
  {Winning arguments: Interaction dynamics and persuasion strategies in
  good-faith online discussions}.
\newblock In \emph{Proceedings of the 25th international conference on world
  wide web}, pages 613--624.

\bibitem[{Taubenfeld et~al.(2024)Taubenfeld, Dover, Reichart, and
  Goldstein}]{Taubenfeld:24}
Amir Taubenfeld, Yaniv Dover, Roi Reichart, and Ariel Goldstein. 2024.
\newblock Systematic biases in llm simulations of debates.
\newblock In \emph{Proceedings of the 2024 Conference on Empirical Methods in
  Natural Language Processing}, pages 251--267.

\bibitem[{Tesauro(1991)}]{tesauro1991practical}
Gerald Tesauro. 1991.
\newblock Practical issues in temporal difference learning.
\newblock \emph{Advances in neural information processing systems}, 4.

\bibitem[{Vacaro et~al.(2019)Vacaro, Marques, Oliveira, Paz, Paula, Staehler,
  and Murphy}]{vacaro2019sim}
Juliano Vacaro, Guilherme Marques, Bruna Oliveira, Gabriel Paz, Thomas Paula,
  Wagston Staehler, and David Murphy. 2019.
\newblock Sim-to-real in reinforcement learning for everyone.
\newblock In \emph{2019 Latin American Robotics Symposium (LARS), 2019
  Brazilian Symposium on Robotics (SBR) and 2019 Workshop on Robotics in
  Education (WRE)}, pages 305--310. IEEE.

\bibitem[{Vaswani et~al.(2017)Vaswani, Shazeer, Parmar, Uszkoreit, Jones,
  Gomez, Kaiser, and Polosukhin}]{vaswani2017attention}
Ashish Vaswani, Noam Shazeer, Niki Parmar, Jakob Uszkoreit, Llion Jones,
  Aidan~N Gomez, Lukasz Kaiser, and Illia Polosukhin. 2017.
\newblock \href {https://openreview.net/forum?id=H1ZGYPb_ZS} {Attention is all
  you need}.
\newblock In \emph{Proceedings of the 31st International Conference on Neural
  Information Processing Systems}.

\bibitem[{Wang et~al.(2019)Wang, Shi, Kim, Oh, Yang, Zhang, and
  Yu}]{wang_persuasion_2019}
Xuewei Wang, Weiyan Shi, Richard Kim, Yoojung Oh, Sijia Yang, Jingwen Zhang,
  and Zhou Yu. 2019.
\newblock \href {https://doi.org/10.18653/v1/P19-1566} {Persuasion for {Good}:
  {Towards} a {Personalized} {Persuasive} {Dialogue} {System} for {Social}
  {Good}}.
\newblock In \emph{Proceedings of the 57th {Annual} {Meeting} of the
  {Association} for {Computational} {Linguistics}}, pages 5635--5649, Florence,
  Italy. Association for Computational Linguistics.

\bibitem[{Xi et~al.(2025)Xi, Chen, Guo, He, Ding, Hong, Zhang, Wang, Jin, Zhou
  et~al.}]{xi2023rise}
Zhiheng Xi, Wenxiang Chen, Xin Guo, Wei He, Yiwen Ding, Boyang Hong, Ming
  Zhang, Junzhe Wang, Senjie Jin, Enyu Zhou, et~al. 2025.
\newblock The rise and potential of large language model based agents: A
  survey.
\newblock \emph{Science China Information Sciences}, 68(2):121101.

\bibitem[{Yang et~al.(2019{\natexlab{a}})Yang, Chen, Yang, Jurafsky, and
  Hovy}]{yang2019let}
Diyi Yang, Jiaao Chen, Zichao Yang, Dan Jurafsky, and Eduard Hovy.
  2019{\natexlab{a}}.
\newblock Let’s make your request more persuasive: Modeling persuasive
  strategies via semi-supervised neural nets on crowdfunding platforms.
\newblock In \emph{Proceedings of the 2019 Conference of the North American
  Chapter of the Association for Computational Linguistics: Human Language
  Technologies, Volume 1 (Long and Short Papers)}, pages 3620--3630.

\bibitem[{Yang et~al.(2019{\natexlab{b}})Yang, Wang, Zhang, Shou, and
  Xu}]{yang2019recurrent}
Ze~Yang, Pengfei Wang, Lei Zhang, Linjun Shou, and Wenwen Xu.
  2019{\natexlab{b}}.
\newblock \href
  {https://www.researchgate.net/profile/Yang-Ze-4/publication/335699354_A_Recurrent_Attention_Network_for_Judgment_Prediction/links/5dc408d9a6fdcc2d2ff858ae/A-Recurrent-Attention-Network-for-Judgment-Prediction.pdf}
  {A recurrent attention network for judgment prediction}.
\newblock In \emph{International Conference on Artificial Neural Networks},
  pages 253--266. Springer.

\bibitem[{Yue et~al.(2018)Yue, Wu, Seshia, Keutzer, and
  Sangiovanni-Vincentelli}]{yue2018lidar}
Xiangyu Yue, Bichen Wu, Sanjit~A Seshia, Kurt Keutzer, and Alberto~L
  Sangiovanni-Vincentelli. 2018.
\newblock A lidar point cloud generator: from a virtual world to autonomous
  driving.
\newblock In \emph{Proceedings of the 2018 ACM on International Conference on
  Multimedia Retrieval}, pages 458--464.

\bibitem[{Zhang and Balog(2020)}]{zhang_evaluating_2020}
Shuo Zhang and Krisztian Balog. 2020.
\newblock \href {https://doi.org/10.1145/3394486.3403202} {Evaluating
  {Conversational} {Recommender} {Systems} via {User} {Simulation}}.
\newblock In \emph{Proceedings of the 26th {ACM} {SIGKDD} {International}
  {Conference} on {Knowledge} {Discovery} \& {Data} {Mining}}, pages
  1512--1520, Virtual Event CA USA. ACM.

\bibitem[{Zhong et~al.(2018)Zhong, Guo, Tu, Xiao, Liu, and
  Sun}]{zhong2018legal}
Haoxi Zhong, Zhipeng Guo, Cunchao Tu, Chaojun Xiao, Zhiyuan Liu, and Maosong
  Sun. 2018.
\newblock \href {https://www.aclweb.org/anthology/D18-1390.pdf} {Legal judgment
  prediction via topological learning}.
\newblock In \emph{Proceedings of the 2018 Conference on Empirical Methods in
  Natural Language Processing}, pages 3540--3549.

\end{thebibliography}

\newpage
\onecolumn

\appendix
\newpage




\section{Expert Strategies}
\label{appendix:strategies}

In this appendix, we present the expert strategies of groups $E_\mathcal{A}$ (in Figure \ref{tab:train_strategies})  and $E_\mathcal{B}$ (in Figure \ref{tab:test_strategies}) of our game.
The formal mathematical notations of the tree node conditions are provided in Table \ref{bots-strategies}.

{\centering
\begin{figure*}[h]
\centering
\label{tab:GroupXStrategies}
\begin{subfigure}{0.32\textwidth}
\centering
\includegraphics[width=\textwidth]{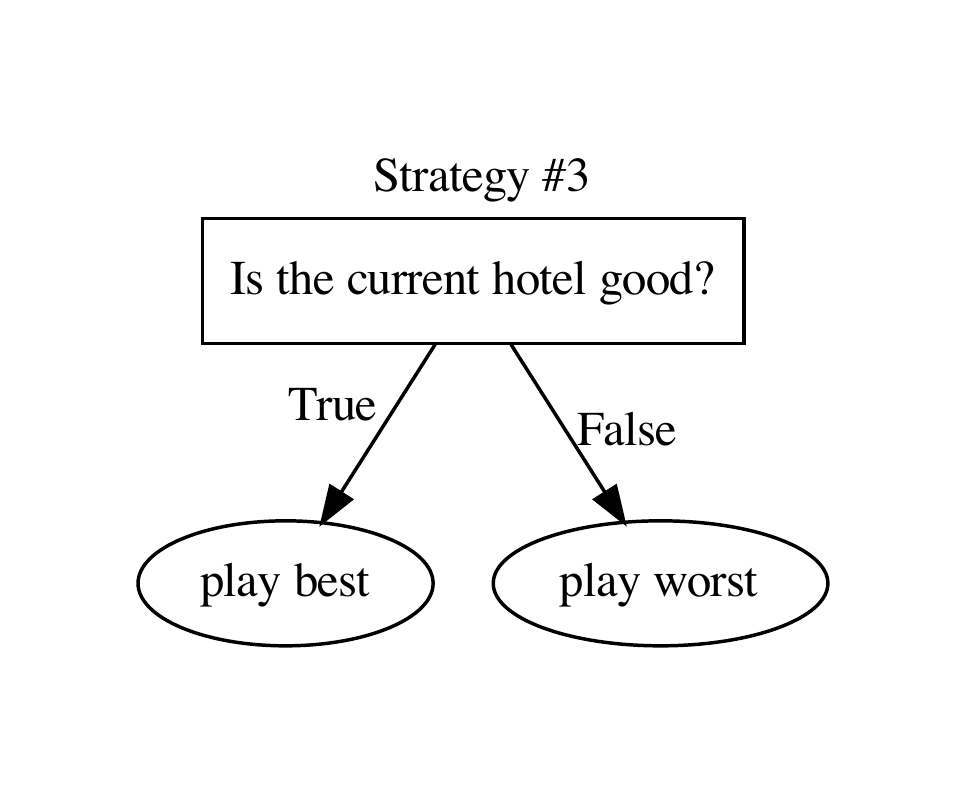}
\caption{1st expert's strategy}
\end{subfigure}
\begin{subfigure}{0.32\textwidth}
\centering
\includegraphics[width=\textwidth]{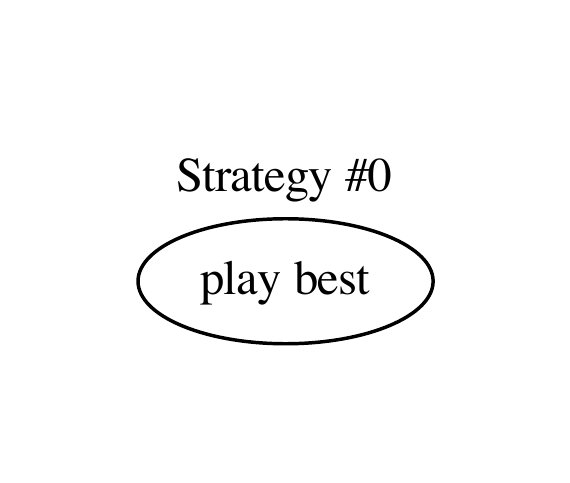}
\caption{2nd expert's strategy}
\end{subfigure}
\begin{subfigure}{0.32\textwidth}
\centering
\includegraphics[width=\textwidth]{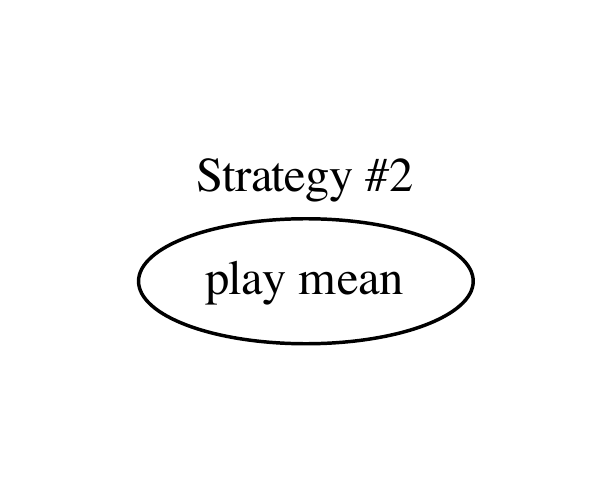}
\caption{3rd expert's strategy}
\end{subfigure}
\begin{subfigure}{0.32\textwidth}
\centering
\includegraphics[width=\textwidth]{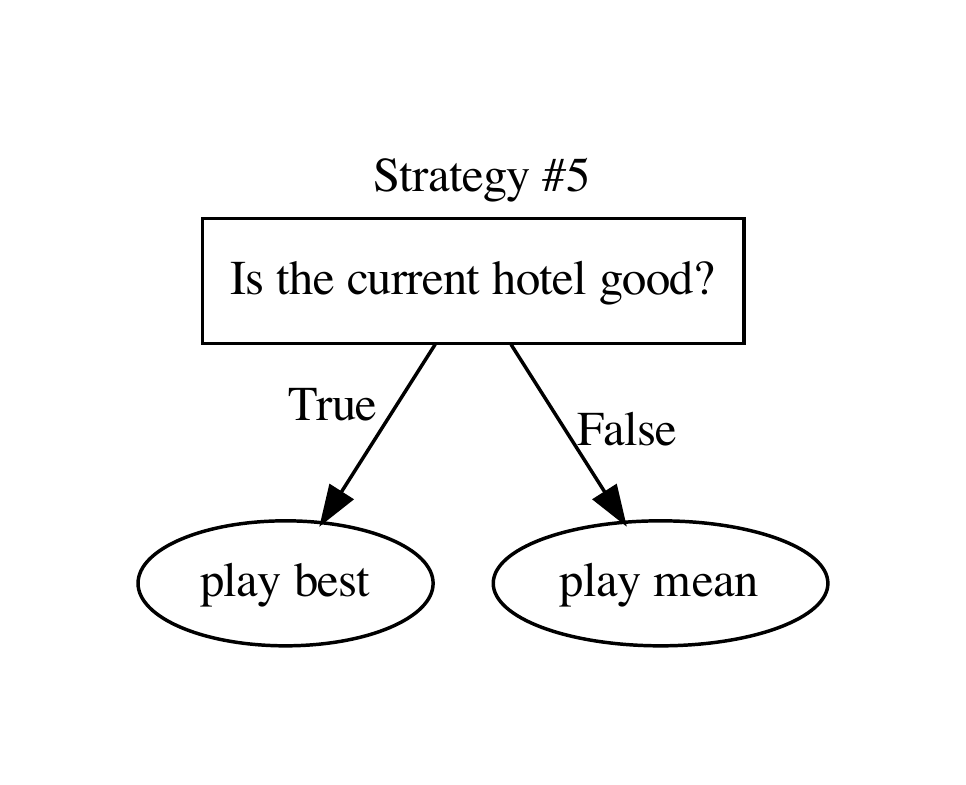}
\caption{4th expert's strategy}
\end{subfigure}
\begin{subfigure}{0.32\textwidth}
\centering
\includegraphics[width=\textwidth]{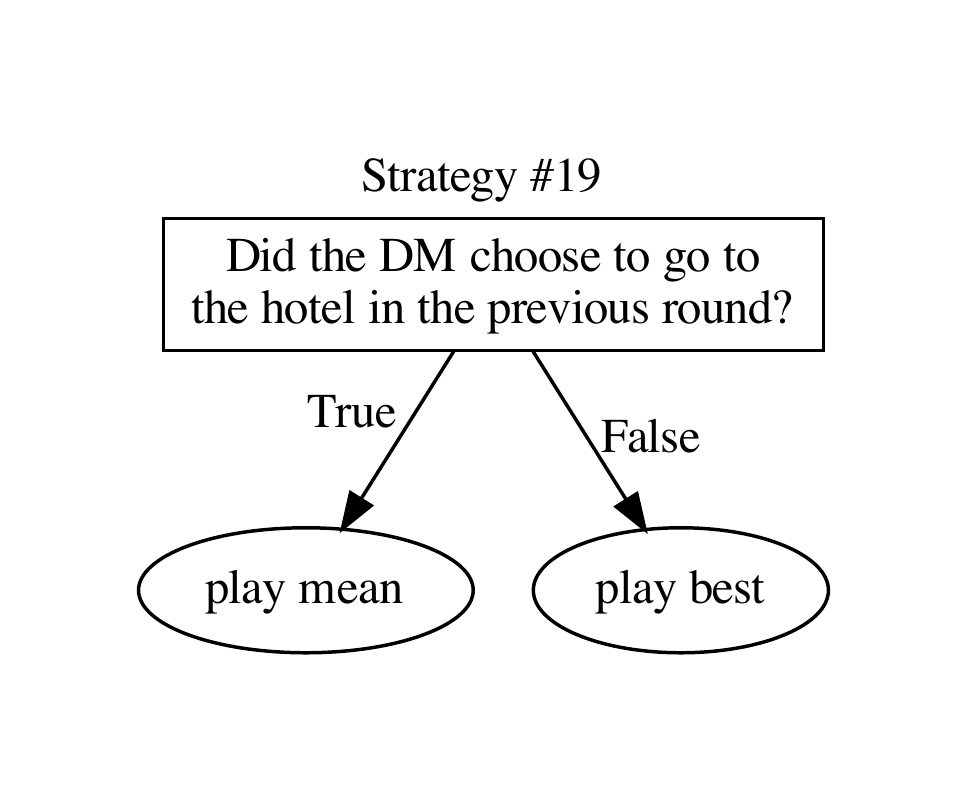}
\caption{5th expert's strategy}
\end{subfigure}
\begin{subfigure}{0.32\textwidth}
\centering
\includegraphics[width=\textwidth]{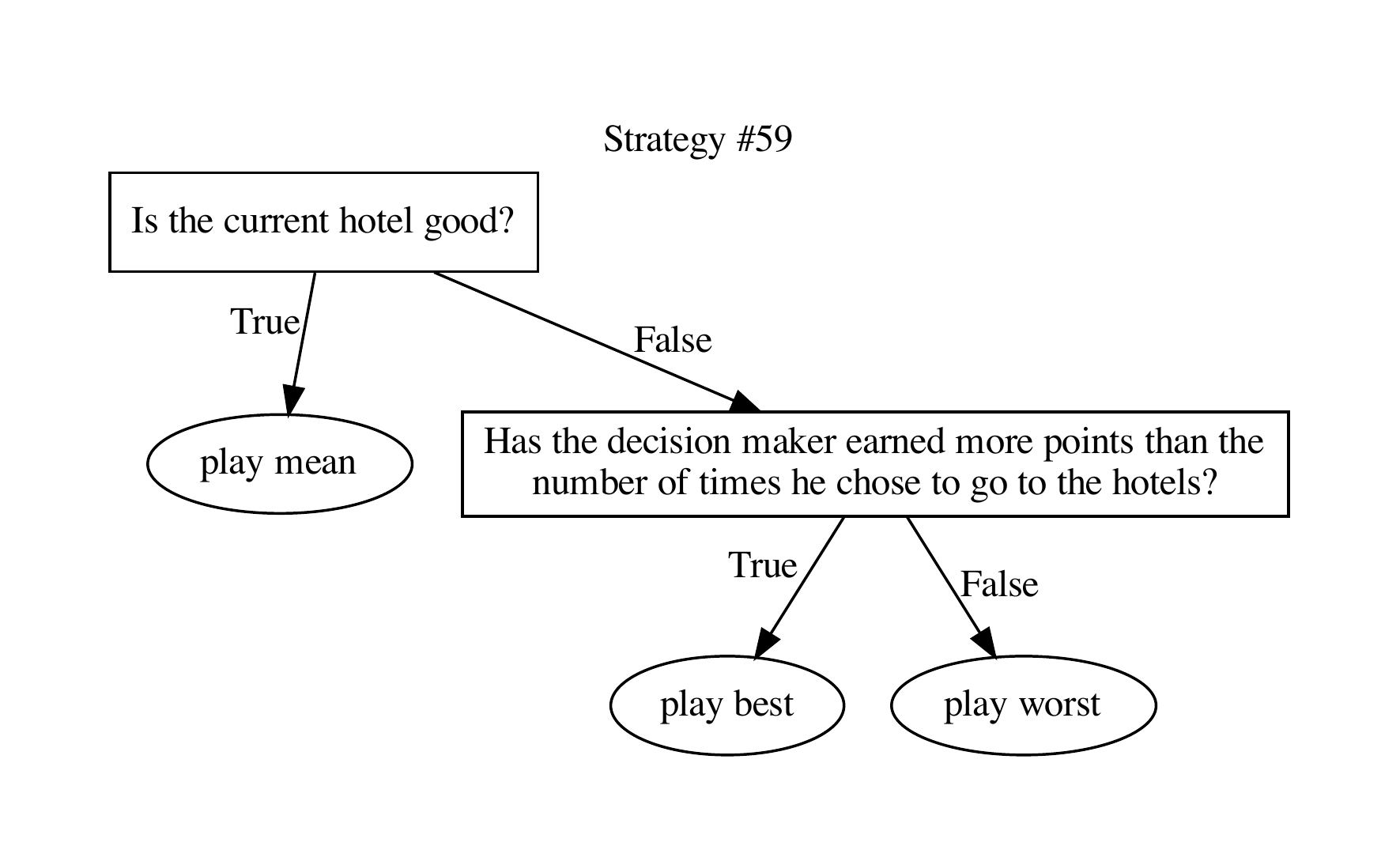}
\caption{6th expert's strategy}
\end{subfigure}
\caption{The strategies of the $E_\mathcal{A}$ experts.}
\label{tab:train_strategies}
\end{figure*}

\begin{figure*}[h]
\centering
\label{tab:GroupYStrategies}
\begin{subfigure}{0.32\textwidth}
\centering
\includegraphics[width=\textwidth]{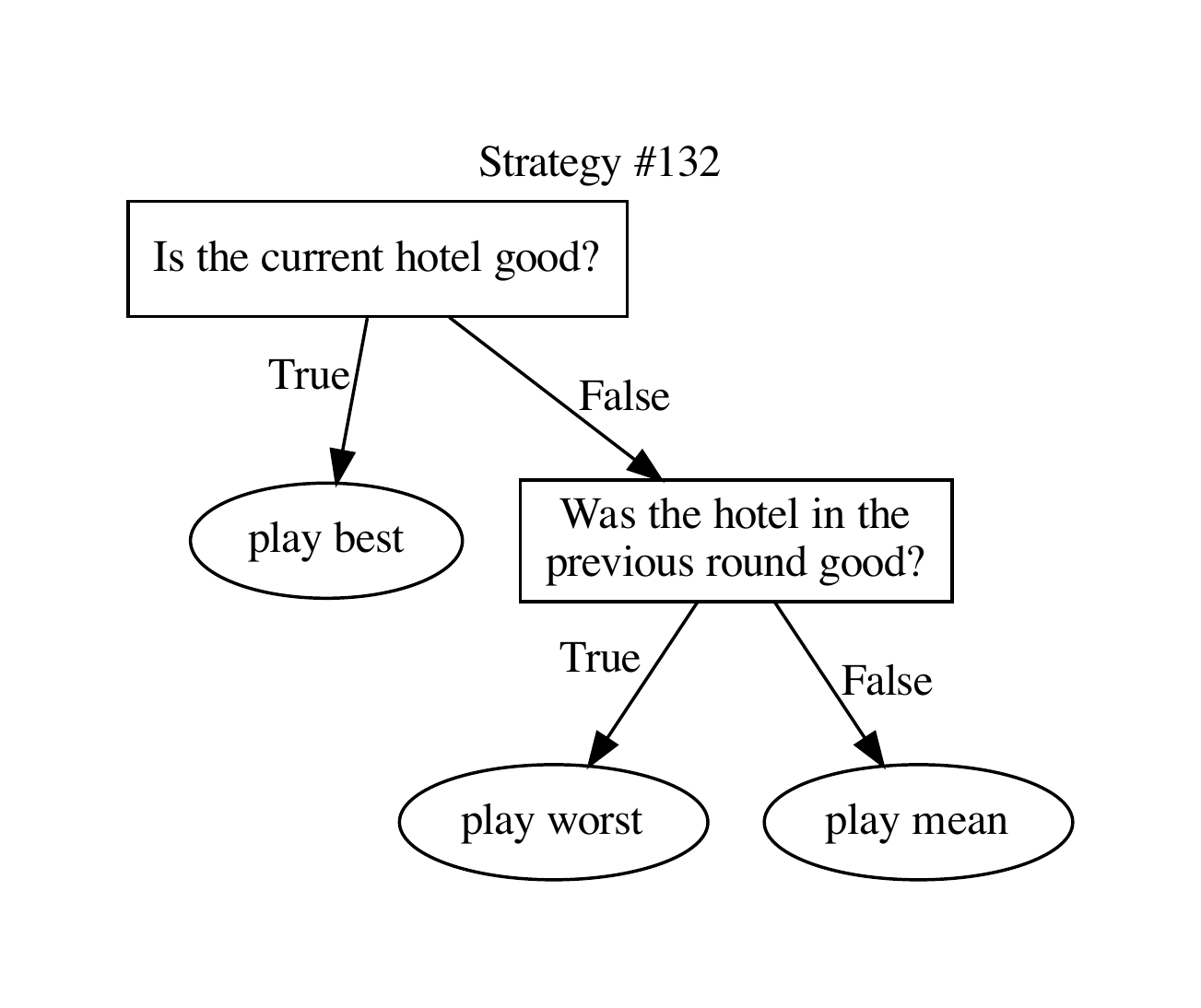}
\caption{1st expert's strategy}
\end{subfigure}
\begin{subfigure}{0.32\textwidth}
\centering
\includegraphics[width=\textwidth]{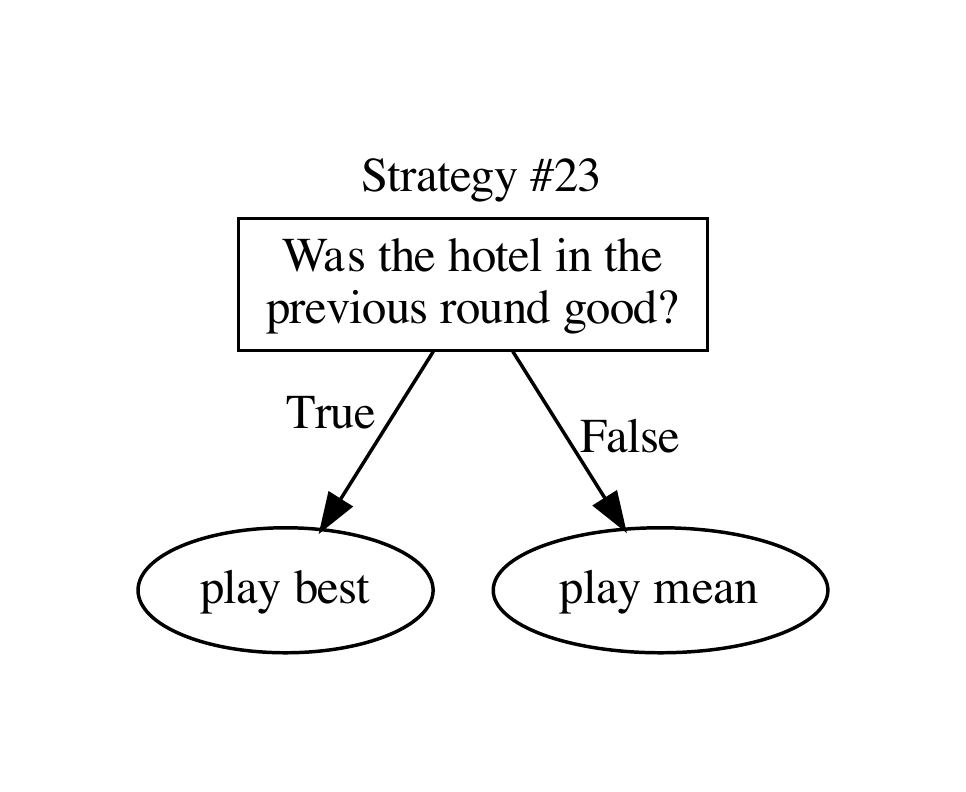}
\caption{2nd expert's strategy}
\end{subfigure}
\begin{subfigure}{0.32\textwidth}
\centering
\includegraphics[width=\textwidth]{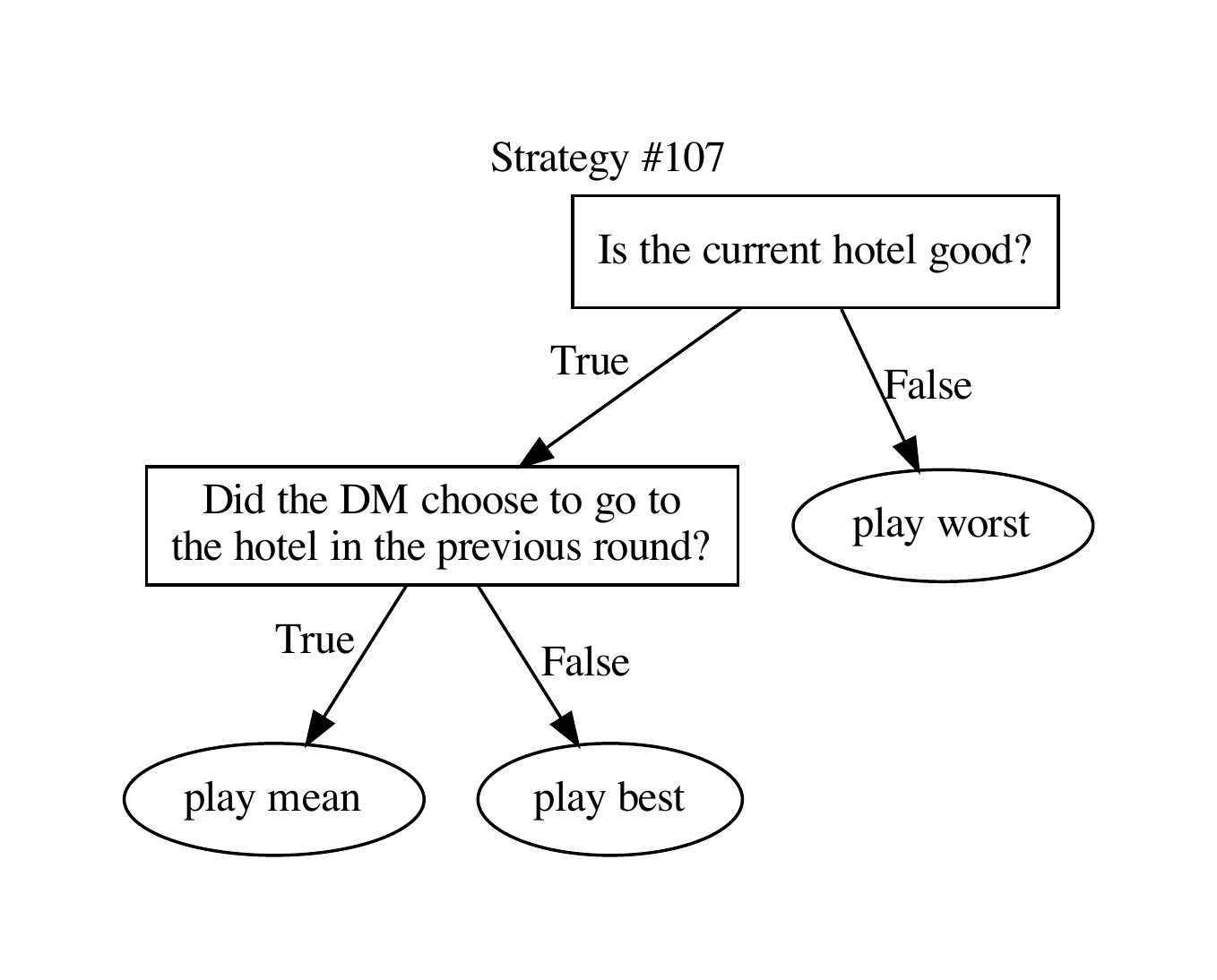}
\caption{3rd expert's strategy}
\end{subfigure}
\begin{subfigure}{0.32\textwidth}
\centering
\includegraphics[width=\textwidth]{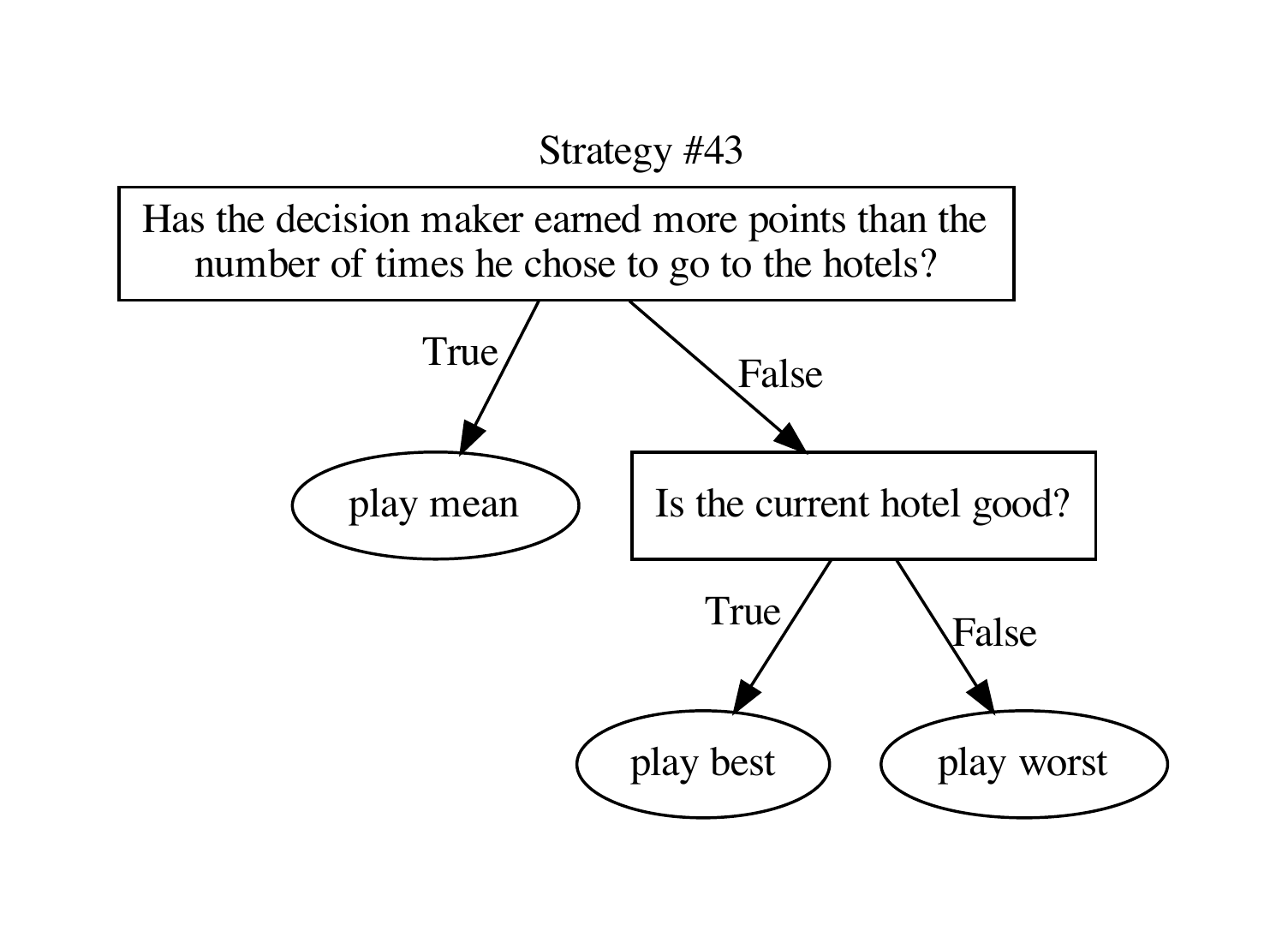}
\caption{4th expert's strategy}
\end{subfigure}
\begin{subfigure}{0.32\textwidth}
\centering
\includegraphics[width=\textwidth]{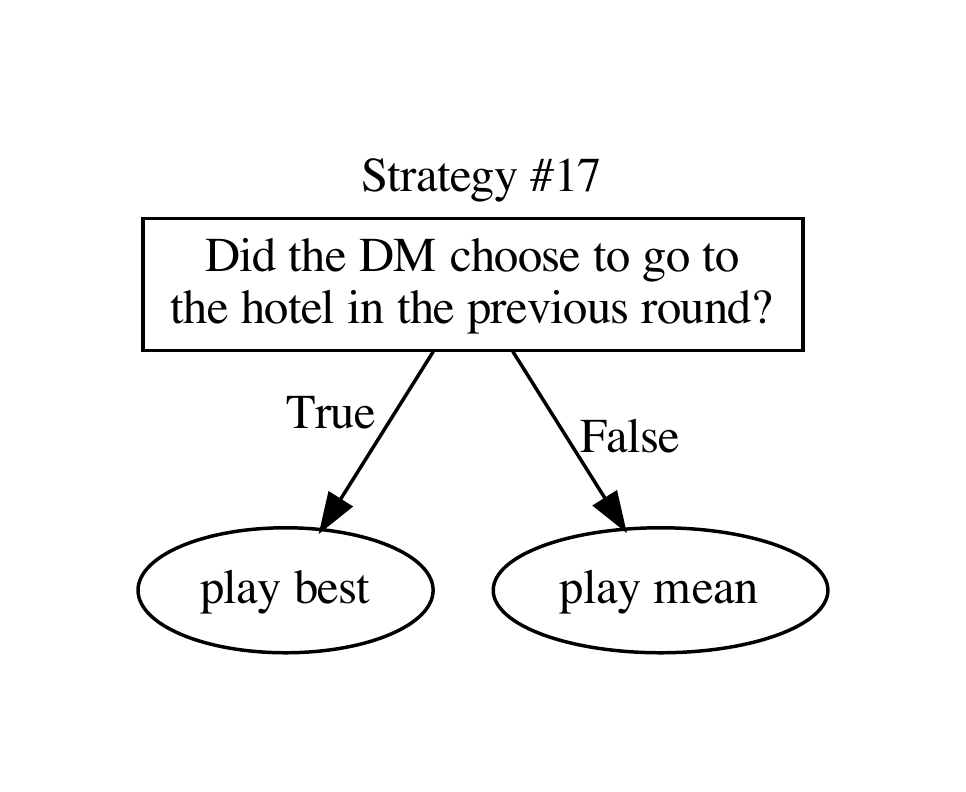}
\caption{5th expert's strategy}
\end{subfigure}
\begin{subfigure}{0.32\textwidth}
\centering
\includegraphics[width=\textwidth]{strategies/93.pdf}
\caption{6th expert's strategy}
\end{subfigure}
\caption{The strategies of the $E_\mathcal{B}$ experts.}
\label{tab:test_strategies}
\end{figure*}
}
\section{Data Collection}
\label{appendix:app}


\subsection{Instructions}
The following text contains the instructions given to players in the app stores.

\qqq{Are you the vacation planner at your house? Think you always know how to choose the best hotel? Start to plan your 10-days trip with our travel agents. Just remember - they don't always want the best for you, and might have their own strategy to make you book the hotel they try to promote!}

\qqq{Travel or Trouble is a strategy game in which you will try to outsmart our traveling agents and plan the perfect vacation for you.}

\qqq{Each game consists of 10 rounds, in each round, one of our traveling agents will introduce you with a review for a new hotel they think might suit you, and you will have to choose: either book the hotel or stay home.}

\qqq{Only true vacation masters can identify a good hotel based upon one review… are you up to the challenge???}

\qqq{As in life, each vacation can turn out to be a great success or a huge disappointment.}

\qqq{Once you made your choice, you will see the results for the vacation in question: was it good or bad?}

\qqq{Based upon the hotel’s average rating (to which only the expert is exposed, and is based on multiple reviews for each hotel), a lottery will determine the outcome of the vacation.}

\qqq{Collect points either by choosing a hotel that turned good or by avoiding bad ones.}

\qqq{Remember - the travel agent is rewarded each time you choose a hotel, regardless of the outcome!}

\qqq{At each game, you will meet a different agent, with a different skill of persuasion.}

\qqq{Try to discover each of our agents’ strategies to persuade you, and take the right decision every round.}

\qqq{Advance through the world of traveling by earning achievements on your way to becoming the true vacation master.}

\subsection{Human Players Information}

As discussed in \S\ref{sec:data_collection}, the app has been available on Google Play and Apple App Store for several months. To attract participants, we also published the app on social media.
To increase participation and game completion (playing until defeating all six experts), in some publications we offered participation in a \$100 lottery for players who completed the game.
In addition, we offered students in an academic course to play the game and complete it in exchange for 0.5 points in the course grade. We therefore know that at least 50\% of the participants are students (as these are the students who received the academic bonus).

\section{The Input of the Models}
\label{appendix:input}

All the models in our experiments utilize the same feature representation. Particularly, each DM-bot interaction round is represented using features relating to both the hotel review shared with the DM, and the strategic situation under which the decision was made. Table \ref{tab:features} illustrates the set of binary Engineered
Features (EFs), a subset of the feature set originally proposed by \citet{apel2020predicting} under the name Hand Crafted Features (HCF),\footnote{As we noted at \S\ref{sec:HCF}, the reason we called the features EFs, while in Apel's work they are called HCFs, is that \citet{apel2020predicting} tagged these features manually, while we labeled them using an LLM.} that are used to represent a review. 
In addition, the table also presents the features we use in order to represent the strategic context of the decision.

\begin{table*}[htbp]
\centering
\small
\label{table:features}
\begin{threeparttable}
\begin{tabular}{lll}
\toprule
\multicolumn{3}{l}{\textbf{Features of the review}}\\
\midrule
\textbf{Category} & \textbf{Feature Description} & \\ 
\midrule
Positive  & Does the positive part of & $t \in$ \{Facilities, Price, Design,\\
Topics & the reviews provide info. about $t$? & Location, Room, Staff, View, \\
&& Transportation, Sanitary Facilities\}
\\
\midrule
Positive Part & Is the positive part empty? & \\
Properties & Is there a positive summary sentence? & \\
 & Number of characters in range $r$? & $r \in$ \{[0,99], [100,199], [200,$\infty$)\}\\
 & Word from group \#$g$\tnote{a} \ in review? & $g \in [1,2,3]$ \\
\midrule
Negative & Does the negative part of & $t \in$ \{Price, Staff, Sanitary\\
Topics & the reviews provide info. about $t$? &  Facilities, Room, Food, \\
& & Location, Facilities, Air\} \\
\midrule
 Negative Part & Is the negative part empty? & \\
 Properties & Is there a negative summary sentence? & \\
  & Number of characters in range $r$? & $r \in$ \{[0,99], [100,199], [200,$\infty$)\}\\
 & Word from group number $g$\tnote{a} \ in review? & $g \in [1,2,3]$ \\
\midrule
Overall  & Is the ratio between the length of  & $r \in$ \{[0, 0.7], (0.7, 4), [4, $\infty$)\} \\ Review  &
the positive part and the negative &  \\
Properties & part\tnote{b} \ in the range r? &\\
\midrule
\multicolumn{3}{l}{\textbf{Features of the 
situation}}\\
\midrule
\textbf{Category} & \textbf{Feature Description} & \\ 
\midrule
 Strategies & Is the previous player action $a$? & $a \in$ \{go, not go\} \\
 Features & Is the previous hotel quality $q$? & $q \in $ \{good, not good\} \\
 & \# of points DM's earned so far. & \\
 & \# of rounds DM's played so far. & \\
 & Points $b$ than rounds played? & $b \in$ \{bigger, not bigger\} \\
 \midrule
 Reaction time\tnote{c} & DM Reaction time in range $r$ seconds? & $r \in$ \{[0, 0.5), [0.5, 1), [1, 2), \\ && [2, 3), [3, 4), [4, 6.5), \\ && [6.5, 12), [12, 20), [20, $\infty$)\} \\
\bottomrule
\end{tabular}
\begin{tablenotes}
\item[a] As described by Apel et al. (2020).
\item[b] In terms of the number of characters.
\item[c] Note that the value of this group of features will be 0 for a simulated DM.
\end{tablenotes}
\end{threeparttable}
\caption{The text-based and strategic features utilized in our work. All features are binary except for the two counting features (denoted with \#).}
\label{tab:features}
\end{table*}

\section{Hyper-parameter Tuning}
\label{appendix:HPT}

\paragraph{Model Architecture}
To identify the most suitable model architecture, we partitioned the DM group that interacts with experts from $E_\mathcal{A}$ into two separate groups. Specifically, we randomly selected 80\% of the DMs to serve as the training group and the remaining 20\% as the validation group. During the training process, we exclusively utilized the interactions between the training group and the first four of the six agents from $E_\mathcal{A}$ as the training data. All the interactions of the validation group DMs, as well as the interactions between the training group DMs and the last two experts from $E_\mathcal{A}$, were employed to assess the model's performance during validation.

To select the hyper-parameters for LSTM (\bestReg{the selected hyper-parameter is underlined}) and LSTM+S (\bestSim{the selected hyper-parameter is colored}), we performed a grid search on the following collection of parameters:
hidden size $\in$ [\bestReg{32}, 64, \bestSim{128}],
learning rate $\in$ [\bestSim{$1e^{-4}$}, $4e^{-4}$, \bestReg{$1e^{-3}$}], and
number of layers $\in$ [\bestRegAndSim{2}, 4, 6]
$S_{r} \in$ [\bestReg{0}, 0.5, 1, 2, \bestSim{4}, 10].

To select the hyper-parameters for Transformer {(\bestReg{the selected hyper-parameter is underlined}) and Transformer+S (\bestSim{the selected hyper-parameter is colored}), we performed a grid search on the following collection of parameters:
hidden size $\in$ [32, \bestSim{64}, \bestReg{128}],
learning rate $\in$ [$1e^{-5}$, $4e^{-5}$, $1e^{-4}$, \bestRegAndSim{$1e^{-3}$}],
number of layers $\in$ [\bestSim{2}, \bestReg{4}], number of heads $\in$ [\bestReg{2}, \bestSim{4}], and
$S_{r} \in$ [\bestReg{0}, 0.5, 1, 2, \bestSim{4}, 10].

To select the hyper-parameters for FC {(\bestReg{the selected hyper-parameter is underlined}) and FC+S (\bestSim{the selected hyper-parameter is colored}), we performed a grid search on the following collection of parameters:
hidden size $\in$ [16, 32, \bestReg{64}, \bestSim{128}],
learning rate $\in$ [\bestReg{$1e^{-4}$}, \bestSim{$4e^{-4}$}, $1e^{-3}$], 
number of layers $\in$ [\bestSim{2}, \bestReg{4}], and
$S_{r} \in$ [\bestReg{0}, 0.5, 1, 2, \bestSim{4}, 10].

To select the hyper-parameters for Mamba {(\bestReg{the selected hyper-parameter is underlined}) and Mamba+S (\bestSim{the selected hyper-parameter is colored}), we performed a grid search on the following collection of parameters:
model dim $\in$ [32, \bestRegAndSim{64}],
state dim $\in$ [\bestReg{32}, \bestSim{64}, 128],
conv dim $\in$ [\bestReg{4}, \bestSim{8}],
learning rate $\in$ [$1e^{-4}$, \bestRegAndSim{$4e^{-4}$}, $1e^{-3}$], and
$S_{r} \in$ [\bestReg{0}, 0.5, 1, 2, \bestSim{4}, 10].

To select the hyper-parameters for XGB {(\bestReg{the selected hyper-parameter is underlined}) and XGB+S (\bestSim{the selected hyper-parameter is colored}), we performed a grid search on the following collection of parameters:
max depth $\in$ [\bestReg{3}, \bestSim{4}, 5, 6],
number of estimators $\in$ [50, \bestSim{100}, \bestReg{250}],
and $S_{r} \in$ [\bestReg{0}, \bestSim{0.5}, 1, 2, 4, 10].

We select the run that achieves the best results while using $S_{r} = 0$ to be the no-simulation version of a given model. 

In all of our experiments we used integer seed values $1 \leq seed \leq 15 $ to reduce noise in architecture selection. We reported the average prediction obtained for the sample for the seed values.




\paragraph{Simulation}

We adjust the hyper-parameters of the simulation model after selecting the appropriate architecture parameters. We tested the $(0, 0.005, 0.1, 0.02)$ values of the $\eta$ DM improvement parameter, where 0 indicates no improvement over time. To avoid an infinite game, we set a cap of 100 games per expert. We tested all possible combinations of $w_1$, $w_2$, and $w_3$ taken from the set (0, 1) to determine the nature vector. To ensure the temperament vector remains normalized, we calculated its components as $(p_1, p_2, p_3) = (\frac{w_1}{\sum_{i=1}^3 w_i}, \frac{w_2}{\sum_{i=1}^3 w_i}, \frac{w_3}{\sum_{i=1}^3 w_i})$. Finally, we assessed the effect of human score estimation noise by varying $\epsilon$, the standard deviation of the normal distribution we used to generate noise, with the values (0.2, 0.3, 0.4).
Based on our experimental results, the best simulation hyper-parameters were found to be $\eta = 0.01$, $(p_1, p_2, p_3) = (\frac{1}{3}, \frac{1}{3}, \frac{1}{3})$, and $\epsilon = 0.3$. We employ these values in our subsequent analysis, as presented in the following section.




\newpage
\section{Additional Experiments and Results}

\subsection{Off- vs. On- Policy Evaluation}
\label{app:ope}

In this appendix, we show that the off-policy task is indeed harder than the on-policy one. 
Given that we have only 35 human players in the test set \( E_\mathcal{B} \), we used leave-one-out cross-validation to assess performance on the on-policy task. For each random seed \( S \) (which controls network randomness), we trained 35 LSTM models: in each iteration, we excluded the data of player \( i \) from \( E_\mathcal{B} \), trained the model on the remaining 34 players, and then used it to label the data for player \( i \). This process allowed us to label all players' data for a given seed \( S \), and we repeated it across 15 different seeds.

To compare with the off-policy task using an identically sized training set (noting that throughout the paper we used the full training set with 210 players), we trained a model on data from 34 randomly selected players from \( E_\mathcal{A} \) for each of the 15 seeds.
The reported results are the average across all seeds, with 95\% confidence intervals.
For off-policy, the model achieved 80.2\% accuracy with a confidence interval of $[79.5, 80.7]$.
For on-policy, the model achieved 81.8\% accuracy with a confidence interval of $[81.6, 81.9]$.
These findings confirm that our off-policy problem is indeed harder, as the on-policy accuracy lower bound is strictly greater than the off-policy accuracy upper bound.

\subsection{Effectiveness of the Simulation in On-Policy Evaluation}
\label{app:onpolicy with sim}

A natural question that arises in the context of the paper is whether the simulation-based approach is effective also in the on-policy scenario. We now show that the answer is positive.
To show this, we repeated the same method as in \ref{app:ope}, but this time, for each seed value, we ran three experiments: the first one without any simulated data, the second one with simulation at \( S_r = 1 \) (i.e., one additional simulated DM for each human DM), and the third one with simulation at \( S_r = 4 \) (i.e., four additional simulated DMs for each human DM). Table \ref{table:onpolicy with sim} shows that adding more simulated DMs indeed improves the accuracy.

\begin{table}[h!]
\centering
\begin{tabular}{|c|c|c|}
\hline
\( S_r \) & \textbf{Mean Accuracy} & \textbf{Accuracy 95\% Confidence Interval} \\
\hline
0 & 81.8 & [81.6, 81.9] \\
1 & 82.0 & [81.9, 82.1] \\
4 & 82.4 & [82.3, 82.5] \\
\hline
\end{tabular}
\caption{Results of the on-policy experiments with different values of \( S_r \). It is evident that the simulation indeed contributes to on-policy evaluation.}
\label{table:onpolicy with sim}
\end{table}

\newpage

\subsection{Models, Baselines, and Hard Examples}

\label{app:main-table}
Table \ref{main-table} presents the accuracy of each of the models, with and without simulated data. Since many examples are either very easy or hard to predict, we focused on the hard examples, as described in \S\ref{chap:results}. 

\begin{table*}[h!]
  \label{accuracy-by-strategy}
  \centering
  \tiny
  \begin{tabular}{|c|c|c|c|c|c|c|c|c|c|c|c|c|}
\hline
\hfill \symbol{92} Hard to & LSTM & LSTM+S & TF & TF+S & Mamba & Mamba+S & FC & FC+S & XGB & XGB+S & Majority & All \\
Accuracy of \symbol{92} & & & & & & & & & & & & Samples\\
\hline
LSTM & 53.7±0.8 & 53.8±0.6 & 58.6±0.6 & 55.7±0.6 & 57.6±0.7 & 52.0±0.5 & 60.2±0.8 & 55.5±0.6 & 59.1±0.6 & 56.6±0.7 & 67.1±0.3 & 82.6±0.1 \\
\hline
LSTM+S & \textbf{60.8±0.8} & \textbf{58.1±0.7} & \textbf{62.8±0.6} & \textbf{61.4±0.6} & \textbf{62.1±0.5} & \textbf{57.5±0.5} & \textbf{63.7±0.7} & \textbf{59.9±0.4} & \textbf{61.2±0.5} & \textbf{61.1±0.5} & \textbf{67.7±0.5} & \textbf{83.6±0.1} \\
\hline
 \hline
TF & 53.9±0.8 & 53.3±0.4 & 56.5±0.8 & 53.7±0.7 & 55.5±0.6 & 50.6±0.6 & 58.7±0.6 & 53.7±0.5 & 57.6±1.1 & 57.0±0.6 & 66.4±0.5 & 82.3±0.1 \\
\hline
TF+S & 59.5±0.7 & 57.6±0.4 & 61.3±0.6 & 58.4±0.6 & 60.4±0.6 & 55.8±0.6 & \textbf{62.8±0.8} & 57.9±0.4 & \textbf{61.4±0.6} & 60.6±0.6 & 66.2±0.5 & 83.4±0.1 \\
\hline
 \hline
Mamba & 55.6±0.8 & 54.2±0.5 & 58.4±0.9 & 55.5±0.9 & 56.2±1.1 & 52.3±0.5 & 60.0±1.1 & 55.8±0.7 & 59.3±0.9 & 57.6±0.6 & 67.1±0.3 & 82.6±0.1 \\
\hline
Mamba+S & \textbf{60.9±0.5} & \textbf{58.9±0.6} & \textbf{62.4±0.4} & \textbf{60.5±0.5} & \textbf{61.8±0.5} & 55.9±0.6 & \textbf{63.9±0.5} & \textbf{59.8±0.5} & \textbf{61.9±0.7} & \textbf{61.8±0.4} & \textbf{67.4±0.5} & \textbf{83.7±0.1} \\
\hline
 \hline
FC & 52.5±1.0 & 52.1±0.6 & 55.8±0.6 & 51.9±0.7 & 53.0±0.7 & 50.0±0.7 & 54.9±1.3 & 50.4±0.8 & 56.3±0.5 & 55.4±0.5 & 64.3±0.5 & 80.9±0.2 \\
\hline
FC+S & \textbf{61.1±0.6} & \textbf{58.5±0.6} & \textbf{61.9±0.6} & 60.0±0.6 & 60.7±0.5 & \textbf{57.4±0.4} & 61.6±0.5 & 56.8±0.6 & \textbf{61.8±0.6} & 59.7±0.6 & 65.7±0.4 & 82.5±0.1 \\
\hline
 \hline
XGB & 55.2 & 53.2 & 56.3 & 55.4 & 56.6 & 52.3 & 56.9 & 55.0 & 49.9 & 51.0 & 65.3 & 81.8 \\
\hline
XGB+S & 57.9 & 57.6 & 58.8 & 59.0 & 60.8 & 56.0 & 61.5 & 58.6 & 57.2 & 57.3 & \textbf{67.8} & 82.9 \\
\hline
 \hline
Majority & 56.1 & 54.8 & 56.4 & 56.2 & 58.4 & 53.9 & 56.2 & 54.0 & 54.9 & 55.1 & 50.5 & 77.4 \\
\hline
 \hline
\# Samples & 2350 & 3743 & 3273 & 3177 & 3246 & 3436 & 2728 & 3959 & 1949 & 2001 & 2709 & 15625 \\ & (15.0\%) & (24.0\%) & (20.9\%) & (20.3\%) & (20.8\%) & (22.0\%) & (17.5\%) & (25.3\%) & (12.5\%) & (12.8\%) & (17.3\%) & (100\%) \\
\hline
\end{tabular}

\caption{Performance of models (rows), with and without simulation, for sets of examples that are challenging for different models (columns), with 95\% confidence intervals. The results demonstrate that training on simulated data improves the accuracy of the models for each subset of challenging examples.} 
\label{main-table}
\end{table*}



\subsection{Contribution of Simulation-based DMs compared to Actual Human DMs}
\label{app: sim vs human contrib}

For this experiment, we denote by $n$ the number of human players available at the outset, and fix \( S_r = 1 \), meaning that we generate another $n$ simulated players when using the simulation. Let $acc(h, s)$ denote the accuracy achieved by training a model with \( h \) human players and \( s \) simulated players. Then, the improvement achieved by the simulation (compared to the initial dataset) is $acc(n,n) - acc(n,0)$, and the improvement achieved by the hypothetical case in which we have access to additional $n$ human players in given by $acc(2n,0) - acc(n,0)$. We refer to the ratio between the two as the \emph{improvement ratio}. Figure \ref{fig:diffrent_train_size} shows the improvement ratio as a function of $n$. It can be seen that our simulation can recover approximately 30\% of the accuracy improvement, at a cost that is effectively negligible (compared to the alternative of doubling the number of human participants).


\begin{figure*}[h!]
  \centering
  \includegraphics[width=0.4\textwidth]{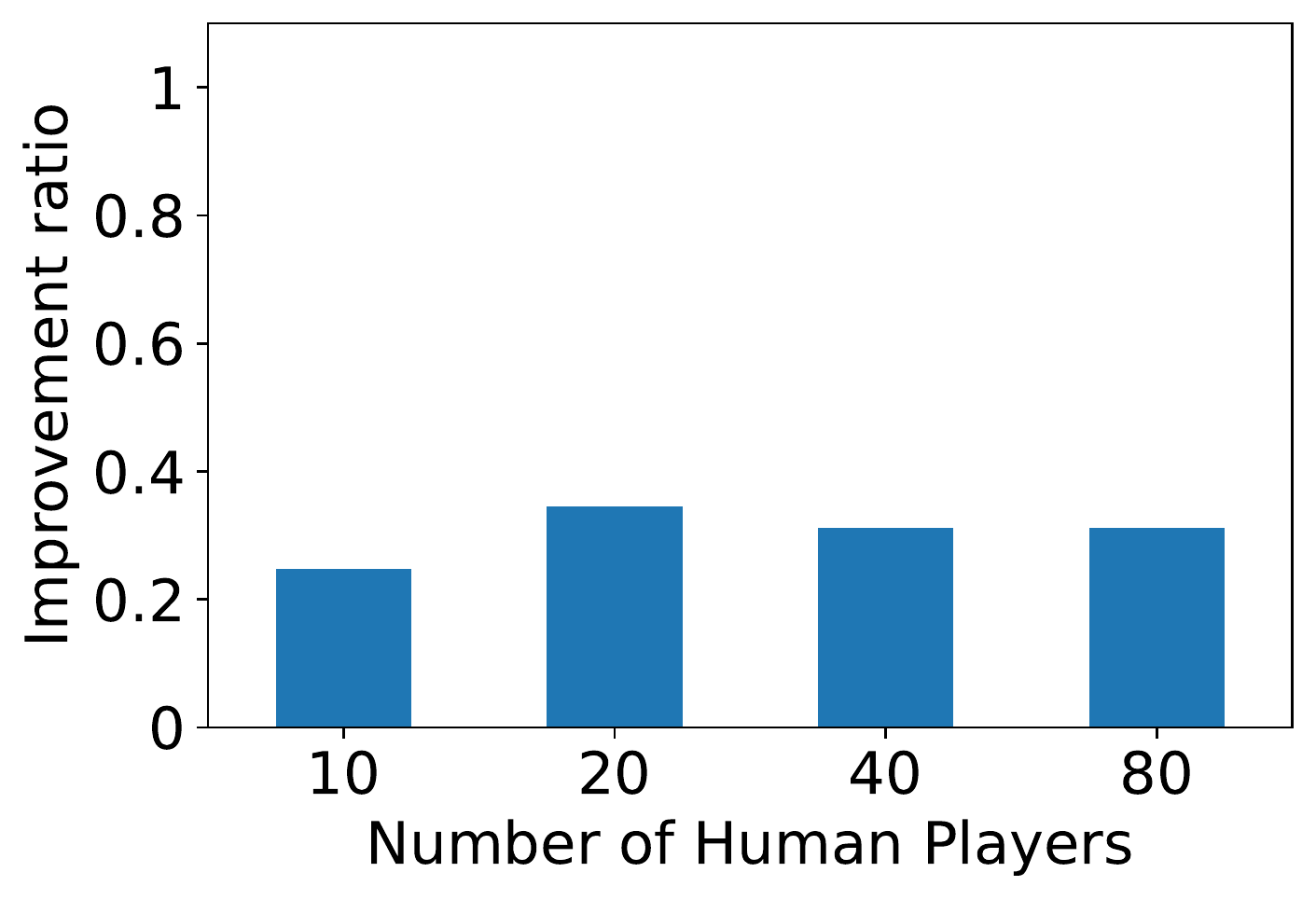}
  \caption{The improvement ratio $\frac{acc(n,n) - acc(n,0)}{acc(2n,0) - acc(n,0)}$ for different values of $n$.}
    \label{fig:diffrent_train_size}
\end{figure*}

\newpage

\subsection{The Impact of Language Representation}
\label{app:lang-representation}

Figure \ref{fig:features} shows the performance of LSTM for the three review representations, as discussed in \S\ref{sec:HCF}: BERT, GPT4 and Engineered Featured (EFs). Notably, as $S_r$ increases, the EF representation outperforms the two alternatives.

\begin{figure}[h!]
  \centering
  \includegraphics[width=0.4\textwidth]{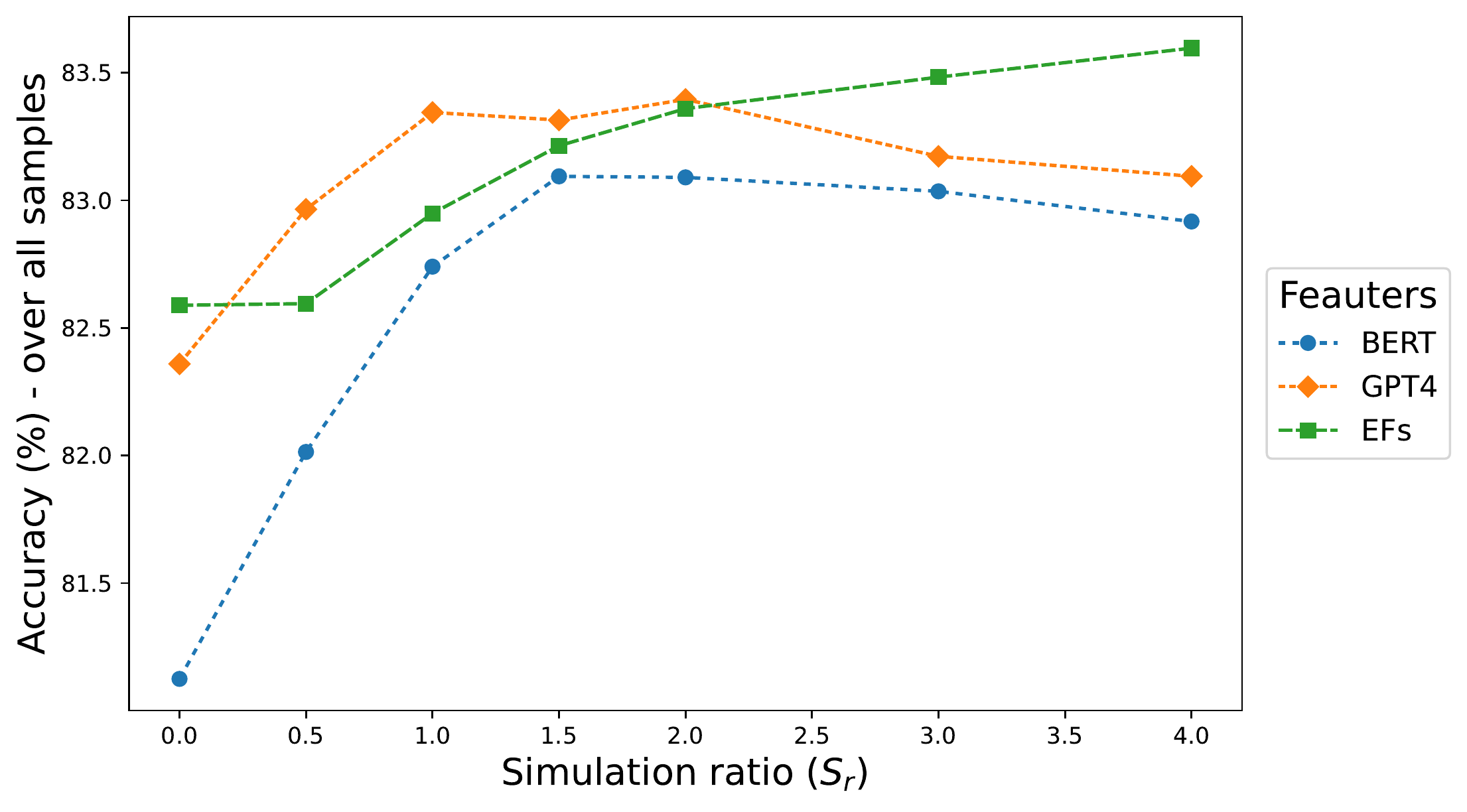}
\caption{The impact of the review representation on the LSTM prediction model.}
    \label{fig:features}
\end{figure}

\end{document}